\documentclass[11pt]{article}

\usepackage[final]{acl}

\usepackage{times}
\usepackage{latexsym}

\usepackage{booktabs}
\usepackage{amsmath, amssymb}

\usepackage[T1]{fontenc}

\usepackage[utf8]{inputenc}

\usepackage{microtype}

\usepackage{inconsolata}

\usepackage{graphicx}

\title{DASH-KV: Accelerating Long-Context LLM Inference via Asymmetric KV Cache Hashing}

\author{
  Jinyu Guo\textsuperscript{1}, 
  Zhihan Zhang\textsuperscript{2},  
  Jiehui Xie\textsuperscript{2}, 
  Md. Tamim Iqbal\textsuperscript{3}, \\
  \textbf{Dongshen Han\textsuperscript{2}, 
  Lik-Hang Lee\textsuperscript{4}, 
  Sung-Ho Bae\textsuperscript{5}, 
  Jie Zou\textsuperscript{2}, 
  Yang Yang\textsuperscript{2}, 
  Chaoning Zhang\textsuperscript{2}\thanks{\ \ Corresponding author.}} 
  \\
  \textsuperscript{1}School of Information and Software Engineering, University of Electronic Science and Technology of China \\
  \textsuperscript{2}School of Computer Science and Engineering, University of Electronic Science and Technology of China \\
  \textsuperscript{3}Bangladesh University of Engineering and Technology \\
  \textsuperscript{4}The Hong Kong Polytechnic University \\
  \textsuperscript{5}Kyung Hee University, School of Computing \\
\texttt{guojinyu@uestc.edu.cn, chaoningzhang1990@gmail.com}
}

\begin{document}

\maketitle
\begin{abstract}
The quadratic computational complexity of the standard attention mechanism constitutes a fundamental bottleneck for large language models in long-context inference. While existing KV cache compression methods alleviate memory pressure, they often sacrifice generation quality and fail to address the high overhead of floating-point arithmetic. This paper introduces DASH‑KV, an innovative acceleration framework that reformulates attention as approximate nearest-neighbor search via asymmetric deep hashing. Under this paradigm, we design an asymmetric encoding architecture that differentially maps queries and keys to account for their distinctions in precision and reuse characteristics. To balance efficiency and accuracy, we further introduce a dynamic mixed-precision mechanism that adaptively retains full-precision computation for critical tokens. Extensive experiments on LongBench demonstrate that DASH‑KV significantly outperforms state-of-the-art baseline methods while matching the performance of full attention, all while reducing inference complexity from \(O(N^{2})\) to linear \(O(N)\).
\end{abstract}

\section{Introduction}

Large-scale pre-trained language models have demonstrated transformative capabilities in tasks including text generation, multi-turn dialogue, and code completion~\cite{achiam2023gpt, touvron2023llama}. However, their deployment in production environments is confronted with significant efficiency challenges. Substantial evidence indicates~\cite{kwon2023efficient, pope2023efficiently} that the computational expense of the standard attention mechanism constitutes a primary performance bottleneck for large language model (LLM) services~\cite{zhou2024survey}. This challenge is especially acute in scenarios demanding sustained context (e.g., multi-turn conversations, long-document generation) .

\begin{figure}[htbp]
  \centering
  \includegraphics[width=\columnwidth]{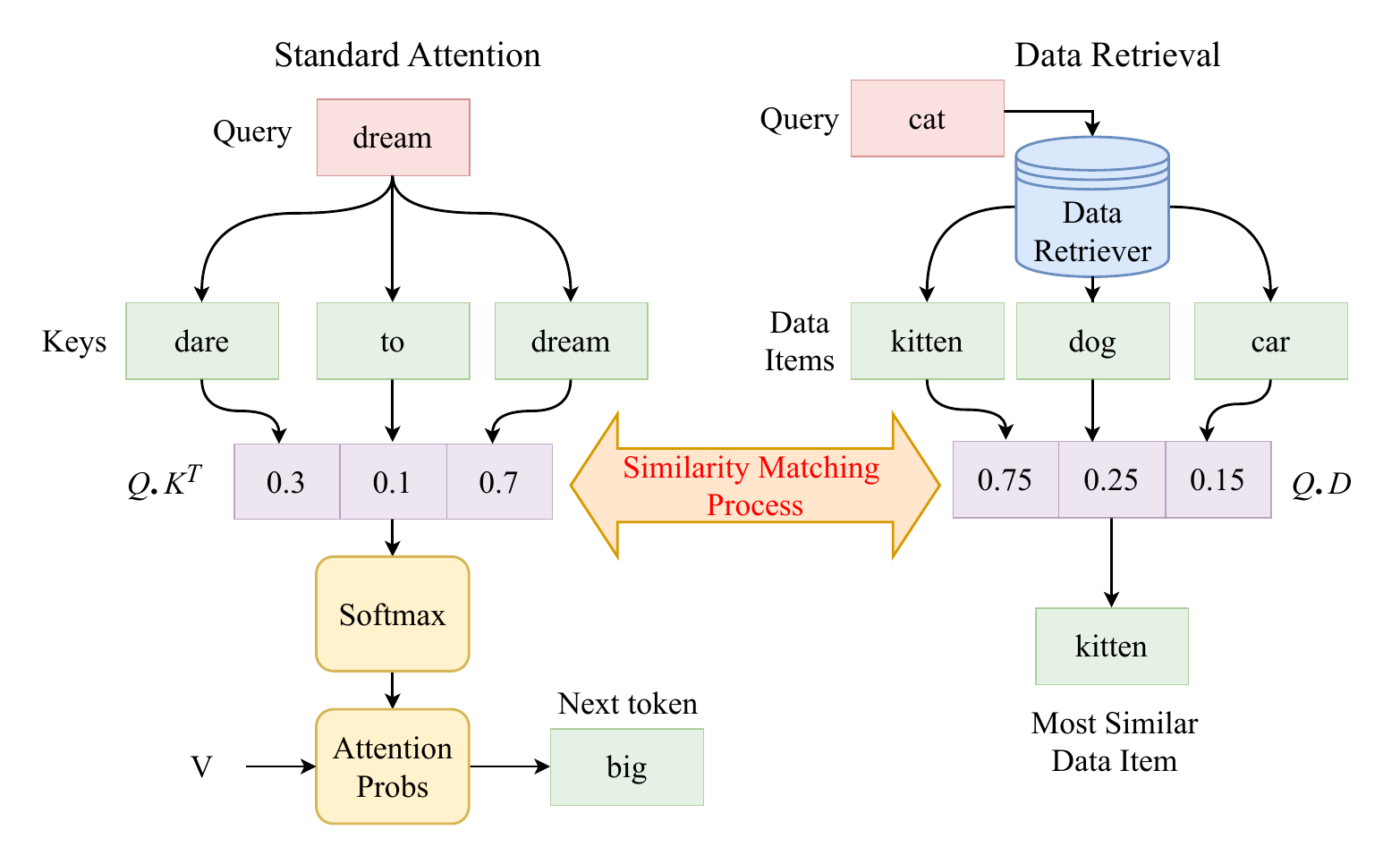}
  \caption{Alignment between Standard Attention Mechanism  and Large-Scale Retrieval}
  \label{fig:teaser}
\end{figure}

To optimize long-text inference, existing research can be broadly categorized into three principal approaches: quantization compression, selective eviction, and structured sharing. Quantization methods ~\cite{liu2024kivi, hooper2024kvquant, zhao2024atom, zheng2026llava} reduce memory footprint by lowering the numerical precision of cached activations. However, employing low bit-widths often precipitates substantial degradation in model performance, especially in ultra-low bit settings (e.g., 1-bit or 2-bit). Furthermore, these techniques typically necessitate de-quantization during computation, incurring additional overhead. Selective eviction strategies ~\cite{xiao2023efficient, zhang2023h2o, li2024snapkv, ge2023model} maintain a fixed-sized cache by retaining only a subset of KV pairs according to specific heuristics. While effective in constraining memory usage, this deterministic eviction results in the permanent loss of discarded information. Such irreversible information loss can impair model efficacy in tasks that rely on long-range dependencies~\cite{zhang2026learning}. Structured sharing techniques ~\cite{vyas2020fast, ainslie2023gqa, kwon2023efficient, dao2022flashattention, sun2026grasp, cao2026language} aim to reduce redundancy by sharing KV parameters across different attention heads or layers. Nonetheless, this approach may not adequately account for the heterogeneous characteristics of different heads or layers and often lacks data-driven adaptability.

More critically, the above static compression strategies do not address the fundamental computational bottleneck of attention mechanisms~\cite{tay2022efficient}. The similarity matching process between queries and keys necessitates billions of high-precision floating-point operations~\cite{dao2022flashattention}. Prevailing optimization methods operate strictly within this floating-point computational framework~\cite{xiao2023smoothquant, yuan2025riemannian, zhang2026lightweight}. Consequently, they leave the underlying computational paradigm unchanged.

Interestingly, the computation of relevance between queries and keys in the attention mechanism is highly similar to relevance-based matching in retrieval tasks. As illustrated in Figure~\ref{fig:teaser}, this perspective reveals a strong alignment between large-scale attention operations and established data retrieval paradigms. In the retrieval task, hashing ~\cite{andoni2015optimal} serves as a prominent method for Approximate Nearest Neighbor (ANN) search, offering advantages in storage efficiency and query speed. Deep hashing ~\cite{liu2012supervised, cao2017hashnet}, in particular, employs neural networks to learn discriminative features and encodes them into compact binary hash codes, thereby enhancing efficiency and reducing computational cost ~\cite{lai2015simultaneous}. Therefore, if this advanced technique from retrieval tasks could be applied to the highly similar attention computation scenarios, there would be significant opportunities for efficiency improvements while maintaining computational accuracy. Furthermore, another supporting point is that prior research indicates that KV caches exhibit low-rank characteristics ~\cite{sharma2023truth, chen2021scatterbrain, kitaev2020reformer}, suggesting they are amenable to compression in a latent space. Consequently, applying deep hashing techniques to the KV cache mechanism presents considerable promise.

In this paper, we propose \textbf{D}eep \textbf{A}symmetric KV Cache Ha\textbf{s}\textbf{h}ing (DASH-KV), which represents the first deep integration of deep hashing techniques into the Transformer attention mechanism to reconstruct the KV Cache computational paradigm. Specifically, our approach transforms the high-dimensional floating-point similarity calculation into a Hamming distance computation between compact binary hash codes. This constitutes a fundamental paradigm shift, replacing costly matrix multiplications with efficient bitwise operations. In addition, to balance precision and efficiency, we propose another two key modules:(1) \textbf{Attention-Oriented Asymmetric Deep Hashing.} Considering that Queries ($Q$) demand higher precision to capture dynamic semantics, whereas Keys ($K$)---which are subject to extensive reuse in the cache---prioritize encoding efficiency, we design an Asymmetric Hashing framework to apply differentiated encoding strategies. Specifically, for $Q$, we employ a deep hashing network (parameterized by a lightweight MLP) to maximize semantic preservation and ensure high-precision retrieval. Conversely, for $K$, we utilize a direct linear projection to facilitate rapid encoding and compact storage, thereby perfectly adapting to scenarios involving massive KV reuse. This differentiated mapping strategy effectively achieves a targeted synergy between computational efficiency and generation quality.
(2) \textbf{Dynamic Importance-Based Mixed-Precision Attention.} Considering the varying importance of different tokens, we propose a lightweight importance predictor, which operates to differentiate between critical tokens (e.g., logical connectors and named entities) and regular tokens. For identified critical tokens, the system switches to full-precision attention. For regular tokens, it employs a high-speed, hash-based attention mechanism. This adaptive paradigm enables fine-grained, instance-level control over the trade-off between computational efficiency and model accuracy.

Experiments on LongBench confirm that DASH-KV outperforms SOTA baselines and achieves performance parity with Full Attention (e.g., on Qwen2-7B). Crucially, it reduces complexity to linear $O(N)$, effectively breaking the efficiency-accuracy bottleneck.

In summary, the principal contributions of this work are as follows:
\begin{enumerate}
\item We propose the DASH-KV framework, which utilizes a novel asymmetric deep hashing paradigm. This architecture is designed to accommodate the distinct roles of queries and keys, enabling the construction of a semantic Hashing space for accelerated inference.
\item We propose an asymmetric deep hashing architecture to address the heterogeneous characteristics of queries and keys. Additionally, we introduce a dynamic mixed-precision mechanism to balance accuracy and efficiency.
\item DASH-KV achieves SOTA and matches Full Attention with linear scaling, breaking the efficiency-accuracy bottleneck.
\end{enumerate}

\section{Related Work}

\subsection{KV Cache Optimization for Long-Context Inference}
Existing research on KV cache optimization mainly pursues three directions: quantization, selective eviction, and structured sharing. Quantization methods (e.g., KIVI, Atom~\cite{liu2024kivi, zhao2024atom}) reduce memory footprint by lowering numerical precision, but they incur accuracy degradation and de-quantization overhead. Selective eviction approaches (e.g., H2O, SnapKV~\cite{zhang2023h2o, li2024snapkv}) maintain a fixed cache size, yet their eviction strategies lead to irreversible information loss. Structured sharing techniques (e.g., GQA~\cite{ainslie2023gqa, sun2026grasp}) aim to reduce parameter redundancy. However, they lack adaptability to the heterogeneous characteristics of different attention heads or layers.

Critically, these methods remain dependent on high-dimensional floating-point computation~\cite{kwon2023efficient}, failing to address the fundamental computational complexity inherent in the attention mechanism.

\subsection{Deep Hashing Methods}
Hashing techniques map high-dimensional vectors to compact binary codes to enable efficient Approximate Nearest Neighbor (ANN) search. The field has progressed from early random projection methods, such as Locality-Sensitive Hashing (LSH)~\cite{andoni2008near}, to learning-based approaches like Iterative Quantization (ITQ)~\cite{guo2016robust}, and further to deep supervised hashing (e.g., DSH~\cite{liu2016deep}) for improved retrieval accuracy.

Despite the success of hashing in large-scale retrieval tasks~\cite{wang2017survey}, its potential synergy with the Transformer attention mechanism has not been extensively investigated~\cite{kitaev2020reformer}. To the best of our knowledge, this work presents the first systematic integration of hashing techniques for attention optimization~\cite{wang2026efficient, wang2026transforming, yuan2026spherical, zhang2026tda}. This is achieved by replacing computationally expensive floating-point similarity operations with efficient bitwise comparisons~\cite{rastegari2016xnor}, thereby accelerating autoregressive inference.

\section{DASH-KV}

\begin{figure*}[t]
\centering
\includegraphics[width=\linewidth, height=0.35\textheight, keepaspectratio]{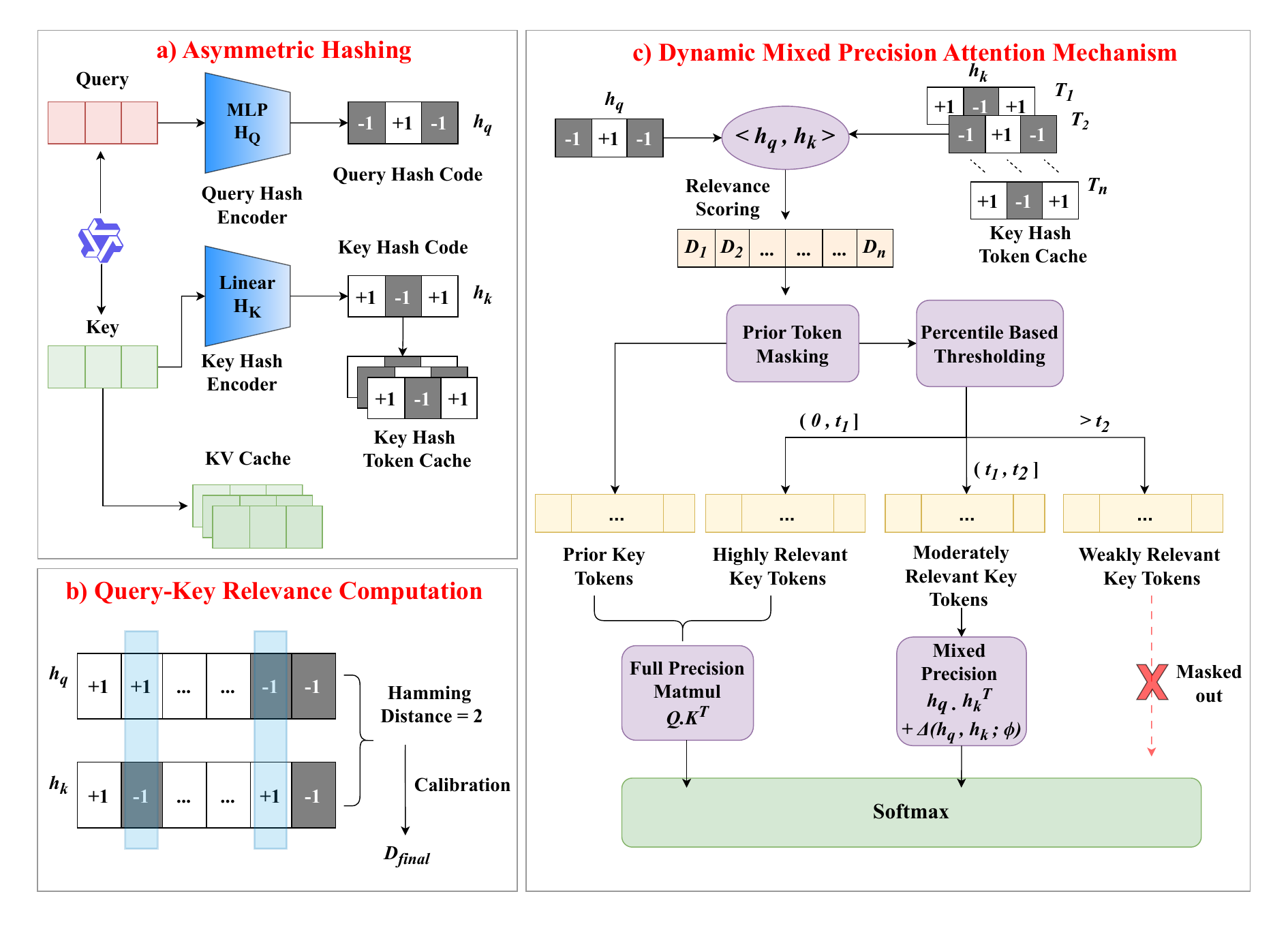}
\caption{Overview of the \textbf{DASH-KV} framework.
It comprises three key stages:
(a) \textbf{Asymmetric Hashing}: Queries are dynamically encoded via a lightweight MLP, while Keys are statically mapped for efficient storage;
(b) \textbf{Query-Key Relevance Computation}: A fast, coarse-grained filtering step using Hamming distance to select relevant candidates;
(c) \textbf{Dynamic Mixed Precision Attention Mechanism}: A dynamic mechanism that retains full precision for critical tokens, applies residual compensation for moderately relevant ones, and masks irrelevant keys from computation, effectively balancing efficiency and accuracy without permanent data loss.}
\label{fig:framework}
\end{figure*}

\subsection{KV Cache Retrieval via Deep Hashing}

We posit that the attention mechanism fundamentally constitutes a process of large-scale vector similarity matching~\cite{kitaev2020reformer, tay2020sparse}. This process exhibits a strong conceptual alignment with core Information Retrieval (IR) tasks, as both are concerned with efficiently identifying the most relevant items from a vast candidate set. This observation motivates us to formally reframe attention computation as a retrieval problem. Consequently, we are inspired to substitute the computationally intensive floating-point dot products with efficient matching in a binary Hamming space. This reformulation establishes a novel and distinct pathway for optimization.

Specifically, we pre-compute and store a binary hash code $h_{k} \in  \{ - 1, + 1\}^l$ for each Key in the KV Cache, where $l$ is the hash code length. During inference, when a new Query arrives, we first encode it into a hash code $h_{q}$. Subsequently, we rapidly identify the subset of Keys most relevant to the Query via Hamming distance. The Hamming distance is defined as:
\begin{equation}
    \text{dist}_{H}(h_{q},h_{k}) = \frac{1}{2}(l - \langle h_{q},h_{k}\rangle),
\end{equation}
where $\langle \cdot , \cdot \rangle$ denotes inner product. A smaller Hamming distance corresponds to a higher similarity.

This paradigm utilizes bitwise operations to achieve hundreds of times acceleration(1-bit vs FP16). However, relying solely on the raw Hamming distance for candidate filtering may introduce bias. To enhance the accuracy of the preliminary retrieval stage, we exploit two inherent architectural properties of Transformers—namely, the multi-headed attention mechanism and its deep layered structure—to design a pair of complementary calibration strategies.

\vspace{1ex}
\noindent\textbf{Cross-Head Consensus.}
The multi-head attention mechanism in Transformers offers multi-view representations of the input~\cite{voita2019analyzing, michel2019sixteen}. Thus, we aggregate the selection preferences for the $i$-th Key across all attention heads by tallying its occurrences:
\begin{equation}
\text{Vote}_{i} = \sum_{h' = 1}^{H} \mathbb{I}(D_{raw}^{(l,h')}(i) < T_{vote}),
\end{equation}
where $H$ is the total number of attention heads, $l$ denotes the current layer, $\mathbb{I}(\cdot)$ is the indicator function, and $T_{vote}$ is a preset voting threshold. If $\text{Vote}_{i}$ is large, it indicates a consensus among heads. In this case, we apply a negative correction to the distance of this Key to increase its importance:
\begin{equation}
\Delta_{spatial}(i) = - \beta_{spatial} \cdot \frac{\text{Vote}_{i}}{H},
\end{equation}
where $\beta_{spatial}$ is a learnable parameter controlling the strength of cross-head consensus. This mechanism mitigates the impact of misjudgments by individual heads through "majority voting."

\vspace{1ex}
\noindent\textbf{Cross-Layer Momentum.}
Relying solely on spatial multi-head information is insufficient. The deep structure of Transformers also contains rich temporal dependencies~\cite{liu2023scissorhands}. Therefore, we introduce a cross-layer momentum mechanism, utilizing the attention distribution of the previous layer as prior information:
\begin{equation}
\Delta_{temporal}(i) = - \gamma_{temporal} \cdot \sigma(A_{i}^{(l-1)}),
\end{equation}
where $A_{i}^{(l-1)}$ is the Attention Score of the $i$-th Key in the previous layer, $\sigma(\cdot)$ is a normalization function (e.g., Sigmoid), and $\gamma_{temporal}$ is a learnable parameter. This mechanism enables the model to inherit "experience" from the previous layer, giving higher priority to continuously attended Keys.

Combining the above two calibration strategies, the final calibrated distance is:
\begin{equation}
D_{final}(i) = D_{raw}(i) + \Delta_{spatial}(i) + \Delta_{temporal}(i).
\end{equation}
By setting $\beta_{spatial}$ and $\gamma_{temporal}$ as learnable parameters, the model can adaptively fuse spatial (multi-head) and temporal (multi-layer) structural priors. This significantly enhances the discriminative power of the hash distance.

\subsection{Asymmetric Hashing }
We propose an Asymmetric Hashing Framework that utilizes independent encoders for queries and keys. 

The \textbf{Query Encoder} is tasked with mapping dynamically evolving query vectors into the hash space, requiring an optimal balance between semantic representational capacity and inference latency. Consequently, we employ a lightweight Multi-Layer Perceptron (MLP) as the backbone architecture. This design avoids the excessive computational overhead associated with more complex models while preserving sufficient expressive power. Specifically, the encoder is implemented as a three-layer fully connected network:
 \begin{equation}
 h_q^{(1)} = \text{GELU}(\text{LayerNorm}(W_1 Q)) ,
 \end{equation} 
 \begin{equation} h_q^{(2)} = \text{GELU}(W_2 h_q^{(1)}),
 \end{equation}
 \begin{equation} v_q = W_3 h_q^{(2)} \in \mathbb{R}^l ,
 \end{equation} 
where $W_1 \in \mathbb{R}^{d \times 256}$, $W_2 \in \mathbb{R}^{256 \times 256}$, and $W_3 \in \mathbb{R}^{256 \times l}$ are learnable parameters.

However, the direct application of the sign function to $v_q$ for binarization results in vanishing gradients, which preclude end-to-end training. Therefore, we employ a Scaled tanh strategy~\cite{cao2017hashnet}. During training, the continuously differentiable tanh function serves as a smooth surrogate for the non-differentiable sign function:
\begin{equation}
\tilde{h}_q = \tanh(\beta \cdot v_q),
\end{equation}
where $\beta$ is a scaling coefficient. Rather than treating it as a fixed hyperparameter, we adopt a dynamic annealing schedule that progressively increases its value throughout the training process:
\begin{equation}
\beta = \min(10.0, 1.0 + \text{global\_step} \times 0.001).
\end{equation}

During inference, we directly apply the sign function to obtain the final binary hash code:
\begin{equation}
h_q = \text{sign}(v_q) \in \{-1, +1\}^l.
\end{equation}

\textbf{Key Encoder.} In contrast to the dynamic generation of queries, keys exhibit high persistence and reusability once written to the KV cache, eliminating the need for repeated encoding. To achieve an optimal trade-off between computational overhead and representational fidelity, we implement a simple yet efficient linear projection for keys. This projection is designed to map the original key vectors into the target hash space with minimal preprocessing latency:
\begin{equation}
z_k = W_k K,
\end{equation}
\begin{equation}
h_k = \text{sign}(z_k) \in \{-1, +1\}^l,
\end{equation}
where $W_k \in \mathbb{R}^{d \times l}$ is a learnable weight matrix.

Thus, the encoding pipeline, which maps raw vectors to compact hash codes and incorporates multi-view calibration for the initial distance estimates, is hereby established.

\subsection{Dynamic Importance-Based Mixed-Precision Attention}

To achieve an optimal balance between precision and efficiency, we formulate a dynamic mixed-precision allocation mechanism that incorporates structural prior knowledge. First, a deterministic retention strategy is enacted through a predefined index set $\mathcal{I}_{prior}$. This ensures the compulsory inclusion of critical tokens, such as [CLS]/[SEP] tokens responsible for global aggregation, sink tokens that preserve numerical stability, and immediate neighbor tokens that capture local dependencies. These essential components bypass the filtering stage and are processed with full-precision attention directly. Subsequently, based on the calibrated distance $D_{final}$, the system adaptively categorizes keys into three distinct tiers, applying full-precision computation, hashing with residual compensation, or computational masking, respectively.

Fixed distance thresholds cannot readily accommodate the distributional variations induced by varying sequence lengths. To address this, we employ an adaptive percentile-based strategy. Specifically, two truncation thresholds are calculated in real-time based on the distribution of $D_{final}$ within the current sequence:
\begin{equation}
    t_{1} = \text{percentile}(D_{final},p_{1}),
\end{equation}
\begin{equation}
    t_{2} = \text{percentile}(D_{final},p_{2}).
\end{equation}

Based on this, we classify Keys into three levels:
\begin{itemize}
    \setlength{\itemsep}{0pt}
    \item \textbf{Highly Relevant ($D \le t_1$)}: Retain full-precision computation $Q \cdot K^T$ to ensure core generation quality is uncompromised.
    \item \textbf{Moderately Relevant  ($t_1 < D \le t_2$)}: Adopt a "Hash + Residual" hybrid scheme to correct quantization errors while balancing precision and efficiency~\cite{jegou2010product}.
    \item \textbf{Weakly Relevant ($D > t_2$)}: Exclude from computation without discarding, maximizing acceleration gains.
\end{itemize}

The treatment of moderately relevant Keys constitutes the core innovation within our mixed-precision mechanism. While these Keys contain non-redundant information, their computational priority does not warrant the expense of full-precision computation. Our strategy, therefore, is to first obtain a rapid approximation using hash codes and subsequently refine this estimate by compensating for quantization errors through a lightweight neural network. Specifically, given hash codes $h_{q},h_{k} \in \{ - 1, + 1\}^{l}$ for Query and Key, we train a residual function $\Delta(h_{q},h_{k};\phi)$ such that:
\begin{equation}
    \text{Attention}(Q, K, V) \approx \text{softmax}(\mathcal{S}) V,
\end{equation}
where the approximate score matrix $\mathcal{S}$ is defined as:
\begin{equation}
    \mathcal{S} = \frac{h_q h_k^T}{d_h} + \Delta(h_q, h_k; \phi).
\end{equation}

Here, $h_{q} \cdot h_{k}^{T}$ provides an efficient rough estimate, while $\Delta$ is responsible for fitting the residual between the hash approximation distribution and the true distribution. Besides, $d_h$ denotes the dimension of the hash codes.

Additionally, we implement a \textbf{Zero Initialization Strategy} for the MLP output layer~\cite{goyal2017accurate} (weights approach 0, bias is 0).

In summary, DASH-KV implements \textbf{fine-grained dynamic allocation}.

\subsection{Loss Design}
Given the intrinsic ranking-based nature of attention, we employ list-wise distillation~\cite{hinton2015distilling} as our primary training objective. We observe that the distribution of normalized hash dot products, constrained to the $[-1, 1]$ interval, is overly smooth and fails to capture the sharper, more peaked distributions characteristic of the original full-precision attention. To address this discrepancy, we depart from the conventional use of symmetric temperature scaling and introduce an asymmetric temperature scaling mechanism to amplify the numerical distinction between relevant and irrelevant keys. The principal loss function is thus defined as the KL divergence between the student and teacher attention distributions:
\begin{equation}
 \mathcal{L}_{\text{distill}} = \text{KL}(P_{\text{student}} \parallel P_{\text{teacher}}). \end{equation}

Furthermore, to ensure the quality of discrete hash codes, we introduce two auxiliary constraints. The \textbf{Bit Balance Loss} aims to maximize information entropy, encouraging the mean of each dimension of the hash code to approach zero~\cite{liu2016deep}:
\begin{equation}
\mathcal{L}_{\text{bal}} = \|\text{mean}(h_q)\| + \|\text{mean}(h_k)\|,
\end{equation}
where $h_q$ and $h_k$ are the binary hash codes for Query and Key, respectively.

The Quantization Loss constrains the hash encoder output to approximate $\{-1, +1\}$~\cite{zhu2016deep, liu2016deep}. This effectively mitigates performance collapse during the transition from continuous representation to discrete encoding in the inference stage:
\begin{equation}
\mathcal{L}_{\text{quant}} = \mathbb{E}\left[(|h_q| - 1)^2\right] + \mathbb{E}\left[(|h_k| - 1)^2\right].
\end{equation}

Combining the above three parts, the final training objective is:
\begin{equation}
\mathcal{L} = \mathcal{L}_{\text{distill}} + \alpha \cdot \mathcal{L}_{\text{bal}} + \beta \cdot \mathcal{L}_{\text{quant}},
\end{equation}
where we set the coefficients $\alpha$ and $\beta$ for the two auxiliary loss terms to $0.1$. This configuration ensures that the dominant gradient signal originates from aligning with the teacher model's distribution. these auxiliary terms impose effective regularization on the quality of the discrete hash codes, preventing convergence to degenerate solutions.

\section{Experiments}

\subsection{Experimental Setup}
\subsubsection{Backbone LLMs}

Our experiments employ three representative, publicly available large language models for comparative evaluation: Qwen2-7B-Instruct, Llama-3.1-8B-Instruct, and Qwen2.5-14B-Instruct. These models encompass diverse architectural designs and parameter scales, enabling us to assess the generalizability of DASH-KV's algorithmic advantages across varied model configurations.

\subsubsection{Tasks}

We evaluate the performance of DASH-KV on six tasks from the LongBench\cite{bai2024longbench} benchmark: NarrativeQA (single-document summarization), HotpotQA (multi-hop reasoning), Qasper (academic paper question answering), MultiNews (multi-document summarization), GovReport (government report comprehension), and TriviaQA (factual question answering). This suite encompasses critical dimensions of long-context understanding, including both single- and multi-document processing, information retrieval, and complex reasoning, as well as factual extraction and abstractive summarization. Collectively, these tasks provide a comprehensive evaluation of the model's generalization ability across diverse long-context scenarios~\cite{zhang2026text}.

\begin{table*}[t]
  \centering
  \small
  \caption{Performance comparison of DASH-KV with baselines on LongBench tasks. We evaluate on three models: Qwen2-7B-Instruct, Llama-3.1-8B-Instruct, and Qwen2.5-14B-Instruct. NQ: NarrativeQA, HQ: HotpotQA, QM: Qasper, MN: MultiNews, GR: GovReport, TQ: TriviaQA.}
  \label{tab:main_results_all}
  
  \resizebox{\textwidth}{!}{
  \begin{tabular}{llccccccc}
    \toprule
    \textbf{Method} & \textbf{Type} & \textbf{NQ} & \textbf{HQ} & \textbf{QM} & \textbf{MN} & \textbf{GR} & \textbf{TQ} & \textbf{Avg} \\
    \midrule
    \multicolumn{9}{c}{\textit{Qwen2-7B-Instruct}} \\
    \midrule
    Full Attn & Dense & 25.13 & 44.04 & 46.13 & 15.42 & 18.06 & 83.47 & 38.71 \\
    StreamLLM & Eviction & 20.47 & 14.31 & 26.97 & 24.88 & 25.70 & 76.56 & 31.48 \\
    H2O & Eviction & 22.88 & 13.30 & 34.28 & 22.72 & 23.69 & 88.75 & 34.27 \\
    SnapKV & Retrieval & 23.86 & 15.60 & 38.61 & 23.07 & 24.56 & 89.31 & 35.84 \\
    \textbf{DASH-KV (Ours)} & Retrieval & 24.65 & 44.50 & 45.34 & 15.40 & 19.13 & 83.33 & 38.73 \\
    \midrule
    \multicolumn{9}{c}{\textit{Llama-3.1-8B-Instruct}} \\
    \midrule
    Full Attn & Dense & 28.11 & 57.43 & 45.29 & 15.20 & 19.97 & 90.22 & 42.70 \\
    StreamLLM & Eviction & 13.60 & 11.70 & 20.10 & 24.70 & 21.50 & 71.10 & 27.12 \\
    H2O & Eviction & 23.10 & 16.00 & 21.30 & 23.40 & 22.30 & 90.20 & 32.72 \\
    SnapKV & Retrieval & 21.30 & 16.60 & 30.80 & 26.30 & 22.20 & 90.20 & 34.57 \\
    \textbf{DASH-KV (Ours)} & Retrieval & 27.61 & 58.01 & 45.35 & 14.90 & 19.25 & 89.47 & 42.43 \\
    \midrule
    \multicolumn{9}{c}{\textit{Qwen2.5-14B-Instruct}} \\
    \midrule
    Full Attn & Dense & 28.20 & 61.98 & 45.52 & 14.19 & 16.78 & 86.91 & 42.26 \\
    StreamLLM & Eviction & 18.30 & 16.40 & 23.40 & 16.04 & 19.01 & 74.10 & 27.87 \\
    H2O & Eviction & 24.40 & 18.00 & 27.60 & 20.99 & 18.96 & 89.70 & 33.28 \\
    SnapKV & Retrieval & 24.10 & 20.00 & 34.60 & 20.90 & 18.97 & 90.00 & 34.76 \\
    \textbf{DASH-KV (Ours)} & Retrieval & 29.52 & 61.83 & 45.79 & 14.44 & 18.06 & 87.88 & 42.92 \\
    \bottomrule
  \end{tabular}
  }
\end{table*}

\subsubsection{Baselines}

To ensure a comprehensive and fair evaluation, we selected four representative baseline methods that cover the major technical approaches for long-context optimization:

\begin{itemize}
  \item \textbf{Full Attention}: Dense attention mechanism, serving as the upper-bound baseline for performance evaluation.
  \item \textbf{StreamingLLM}\cite{xiao2023efficient}: Leverages the attention sink phenomenon to enable smooth inference over infinitely long texts.
  \item \textbf{H2O}\cite{zhang2023h2o}: Dynamically evicts low-contribution KV pairs to maintain a constant-size cache.
  \item \textbf{SnapKV}\cite{li2024snapkv}: Identify the most important prompt features in the "observation window" at the end of the input to compress the LLM KV cache.
\end{itemize}

\begin{table*}[t]
  \centering
  \small
  
  \caption{Efficiency vs. Accuracy trade-off analysis comparing DASH-KV with symmetric hashing and naive LSH baselines. $P_h$ and $P_f$ denote the attention distributions of hash-based approximation and full-precision calculation, respectively.}
  \label{tab:component_ablation}
  
  \begin{tabular}{lccc}
    \toprule
    \textbf{Variant} & \textbf{Recall@100} & $\mathbf{KL}(P_h \parallel P_f)$ & \textbf{Latency per Token} \\
     & \textbf{(\%)} & \textbf{($\downarrow$)} & \textbf{(ms)} \\
    \midrule
    DASH-KV-Naive & 8.91 & 3.2000 & 28 \\
    DASH-KV-Sym & 68.28 & 1.3200 & 38 \\
    \textbf{DASH-KV (Ours)} & \textbf{86.06} & \textbf{0.4054} & \textbf{22} \\
    \bottomrule
  \end{tabular}
\end{table*}

\begin{figure*}[t]
  \centering
  \begin{minipage}[t]{0.32\textwidth}
    \centering
    \includegraphics[width=\linewidth, height=0.16\textheight]{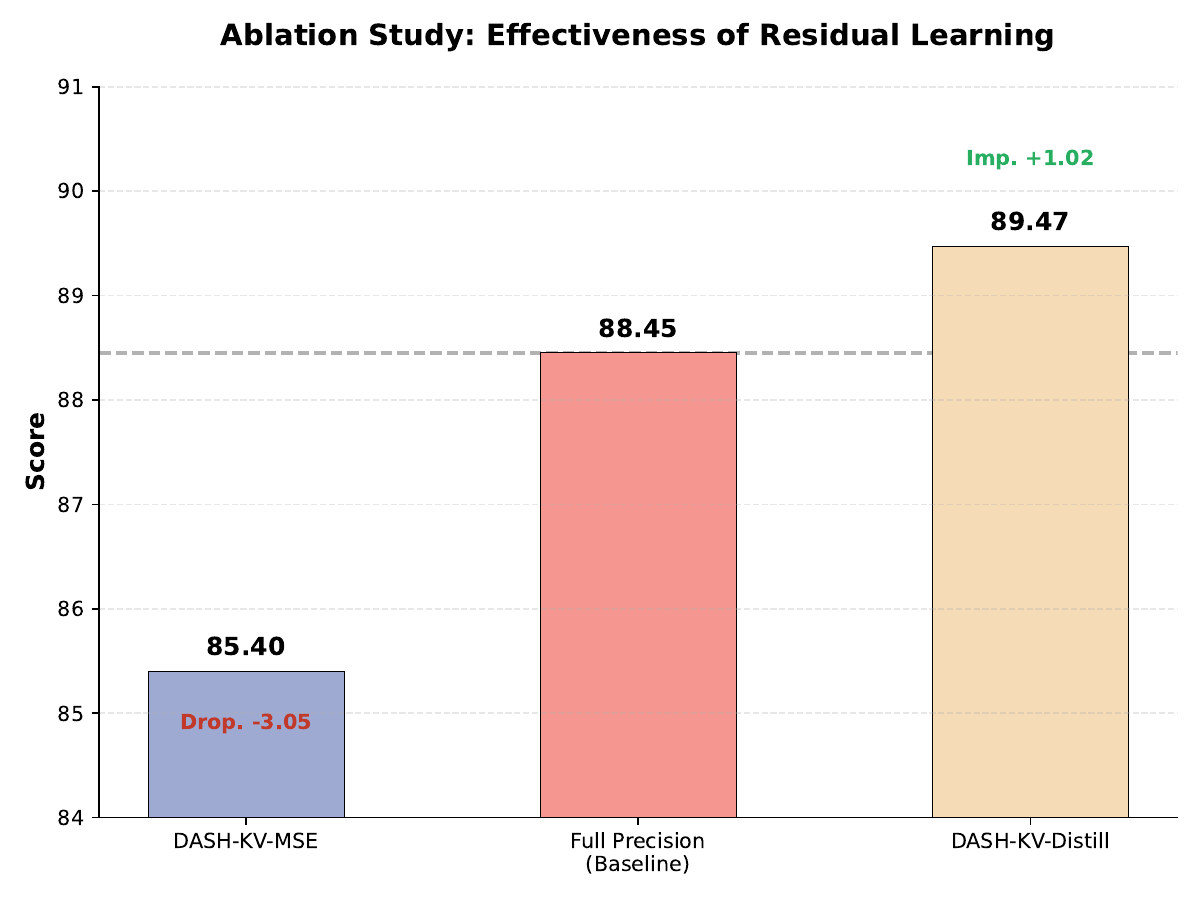}
    \caption{Ablation Study on TriviaQA - F1 Score Comparison}
    \label{fig:ablation_residual}
  \end{minipage}
  \hfill
  \begin{minipage}[t]{0.32\textwidth}
    \centering
    \includegraphics[width=\linewidth, height=0.16\textheight]{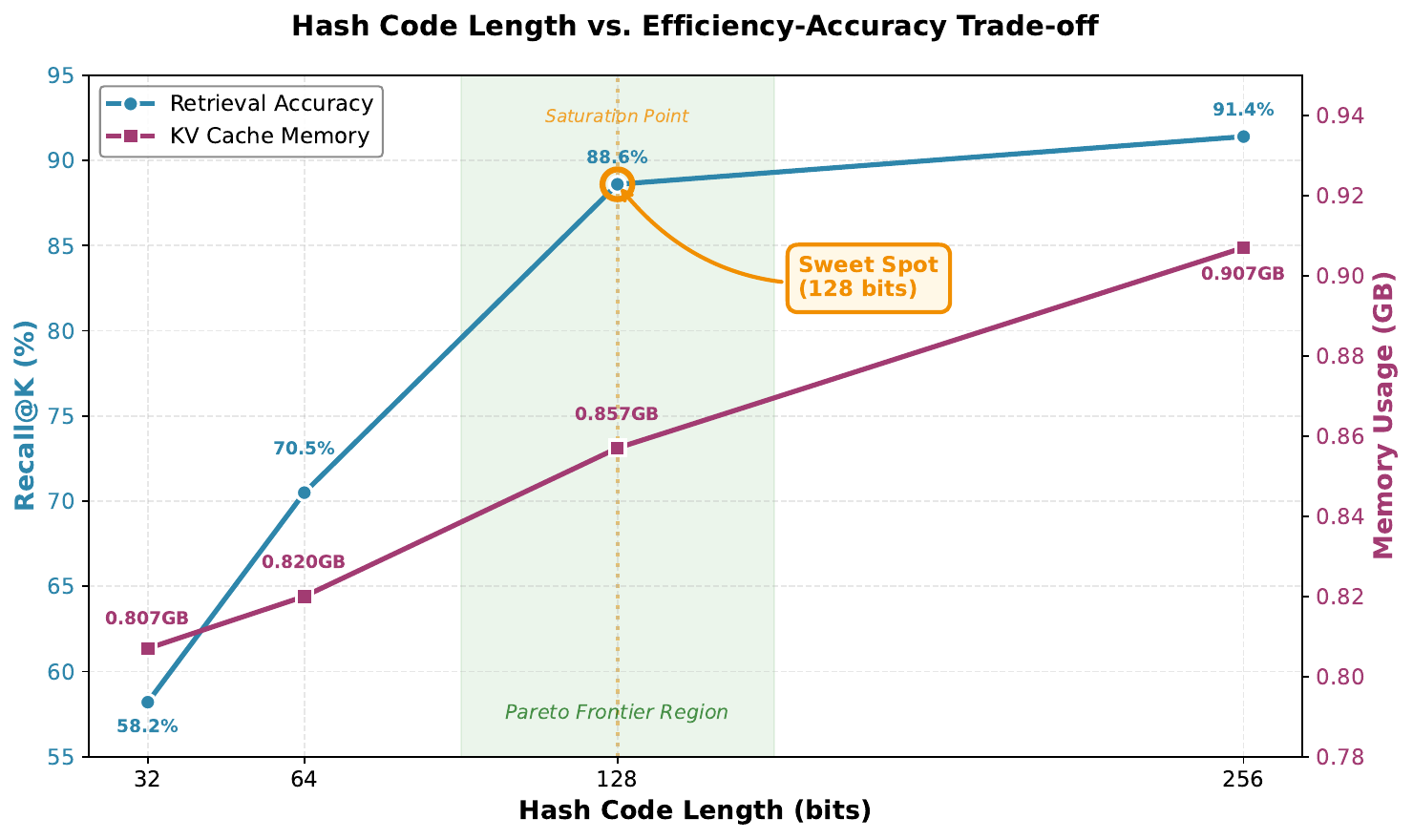}
    \caption{Pareto Frontier of Efficiency and Accuracy}
    \label{fig:pareto_frontier}
  \end{minipage}
  \hfill
  \begin{minipage}[t]{0.32\textwidth}
    \centering
    \includegraphics[width=\linewidth, height=0.16\textheight]{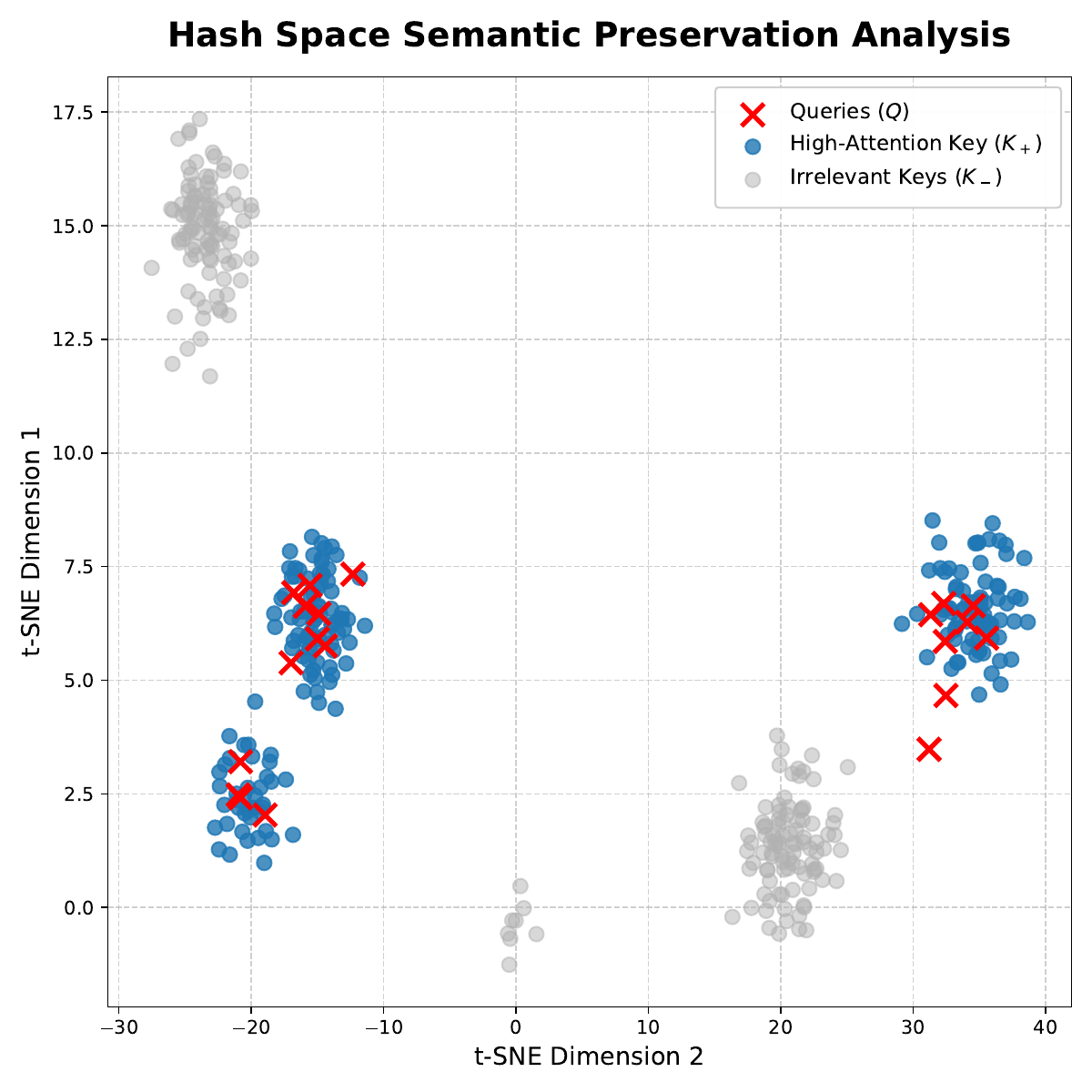}
    \caption{t-SNE Visualization of Hash Codes}
    \label{fig:t-SNE}
  \end{minipage}
\end{figure*}

\subsubsection{Implementation details}

To reflect the prevalent "short-text training, long-text inference" paradigm in practical deployment, we employ a training sequence length of 3k tokens. During inference, the sequence length is extended to 32k tokens. All experiments are conducted under consistent hardware and hyperparameter configurations to ensure a fair and controlled environment. For implementation expediency and to utilize PyTorch's optimized kernels, binary hash codes were simulated using FP16 format during prototyping. It is important to emphasize that all memory footprint statistics reported in this work are calculated based on an efficient 1-bit packed storage model. This model can be directly deployed in production via low-level bitwise operations to achieve the stated compression benefits.

\subsection{Main Results on LongBench Tasks}

The results presented in Table \ref{tab:main_results_all} demonstrate that DASH-KV attains significant inference acceleration while maintaining accuracy. This gain in efficiency arises from the use of Hamming distance computations on 1-bit hashes. This approach provides a decisive computational advantage over the resource-intensive floating-point matrix multiplications required in full-precision models.

\subsection{Ablation Studies}
We conduct a systematic ablation study to validate the contribution of each component within the DASH-KV framework. Specifically, we analyze the role of the Residual Compensation Strategy in mitigating errors introduced by hash quantization.

\subsubsection{Effectiveness of Residual Learning}

To quantify the contribution of residual learning, a comparison is conducted among three system variants: Pure Hash (which operates without residual compensation), DASH-KV-MSE (optimized via Mean Squared Error), and DASH-KV-Distill (trained utilizing Knowledge Distillation).

The results presented in Figure~\ref{fig:ablation_residual} demonstrate a notable phenomenon.

\subsubsection{Component Effectiveness}

The results presented in Table~\ref{tab:component_ablation} demonstrate that the asymmetric design yields a substantial improvement in accuracy. While Locality-Sensitive Hashing (LSH) is computationally efficient due to its parameter-free formulation, its retrieval precision is lower than that of our trained hash encoder, as reflected by a markedly reduced Recall@100. This outcome supports the conclusion that, within the specific context of attention retrieval, a data-driven hash space optimized explicitly for semantic similarity preservation significantly outperforms generic geometric hashing approaches.

\subsubsection{Parameter Sensitivity}

The experimental results in Figure~\ref{fig:pareto_frontier} illustrate the trade-off between hash code length and efficiency.

\section{Analysis}

\subsection{Semantic Preservation in Hamming Space}

Figure~\ref{fig:t-SNE} illustrates the semantic preservation capacity of the proposed hash encoding. Within the learned hash embedding space, semantically related Queries and Keys form compact clusters, indicating cohesive grouping. In contrast, irrelevant Keys are effectively dispersed, resulting in a distinct and significant spatial separation between dissimilar instances.

\subsection{Scalability to Ultra-Long Context}

To assess the scalability of DASH-KV in processing extended-length texts, systematic stress testing was conducted on synthetic sequences spanning 4k to 128k tokens.

\begin{figure*}[t]
  \centering
  \begin{minipage}[t]{0.32\textwidth}
    \centering
    \includegraphics[width=\linewidth, height=0.16\textheight]{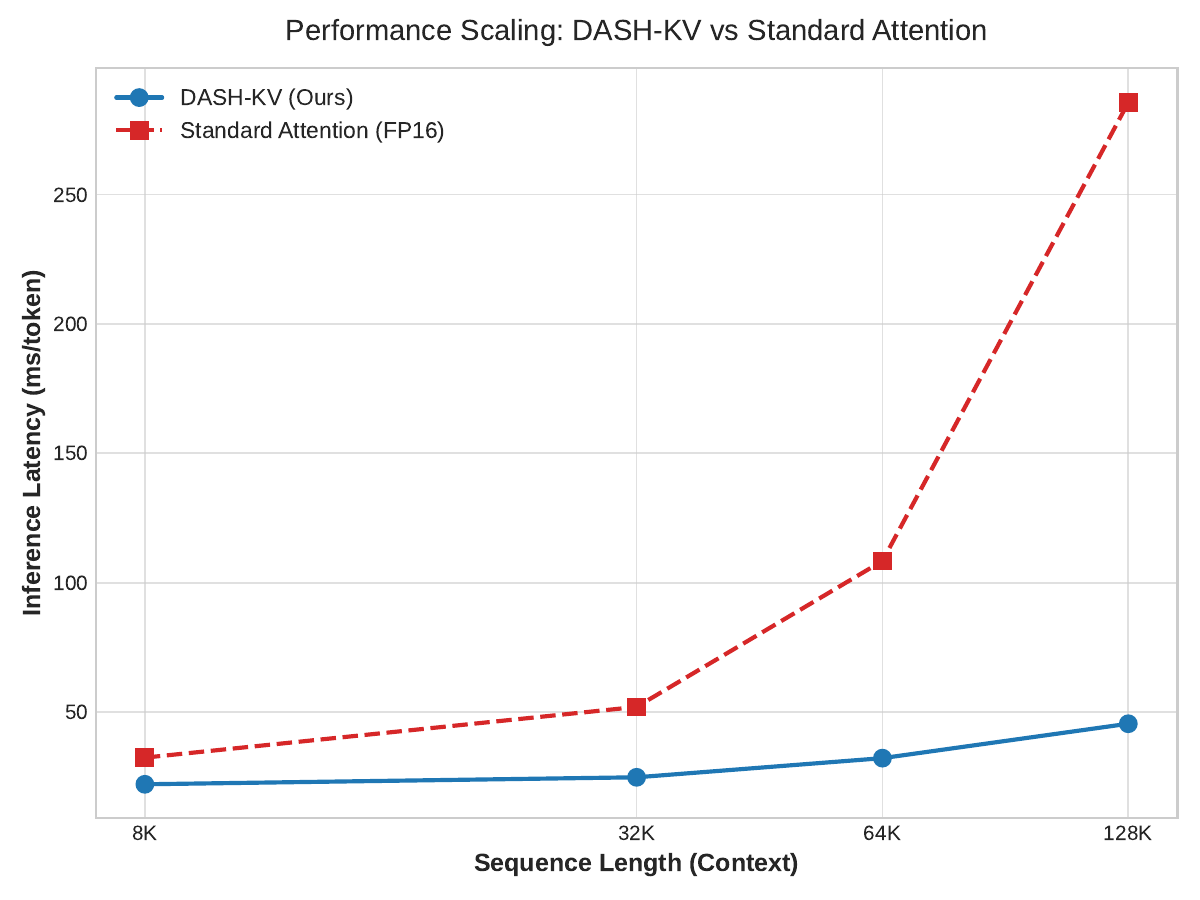}
    \caption{Latency Scaling Curves}
    \label{fig:scalability}
  \end{minipage}
  \hfill
  \begin{minipage}[t]{0.32\textwidth}
    \centering
    \includegraphics[width=\linewidth, height=0.16\textheight]{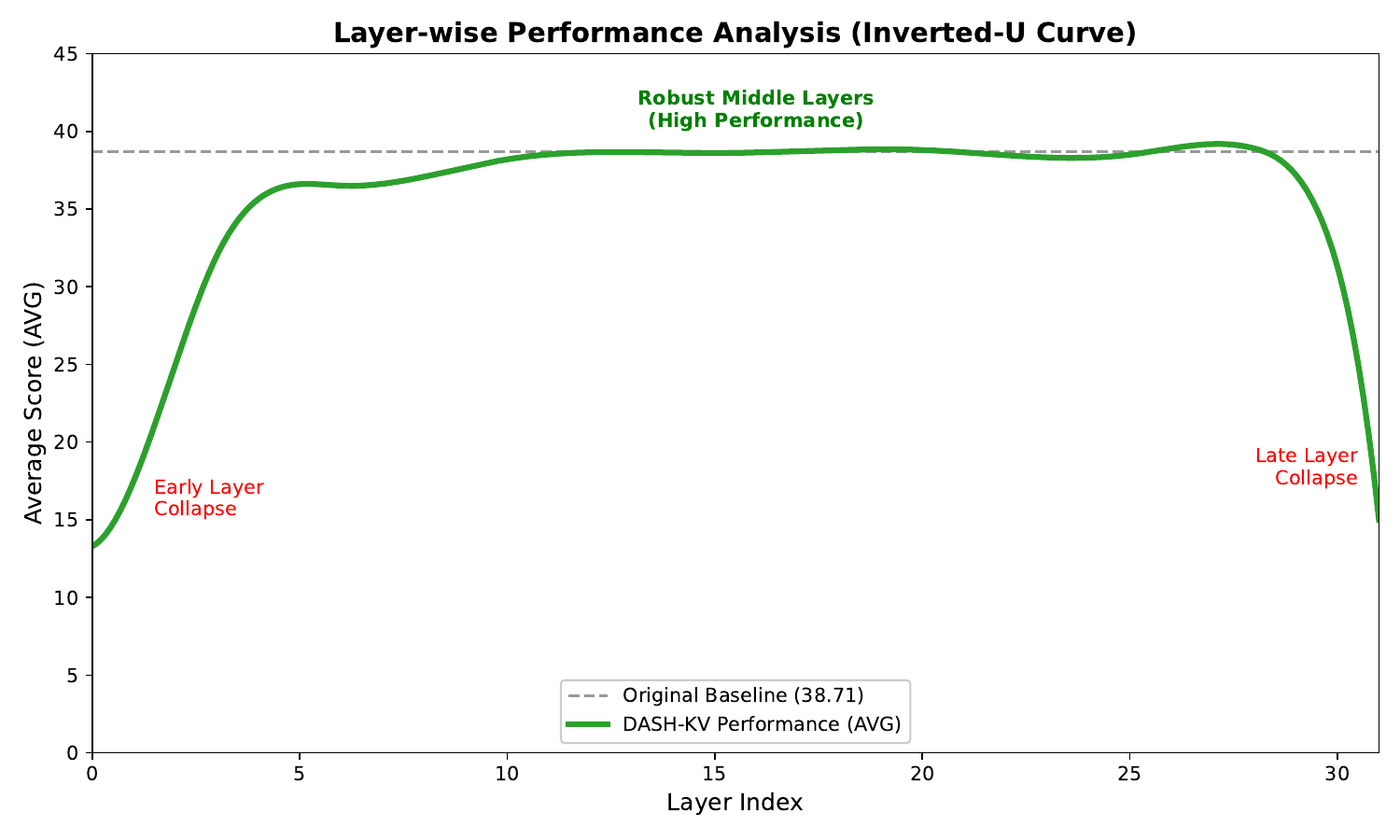}
    \caption{U-shaped Curve of Layer Sensitivity}
    \label{fig:layer_sensitivity}
  \end{minipage}
  \hfill
  \begin{minipage}[t]{0.32\textwidth}
    \centering
    \includegraphics[width=\linewidth, height=0.16\textheight]{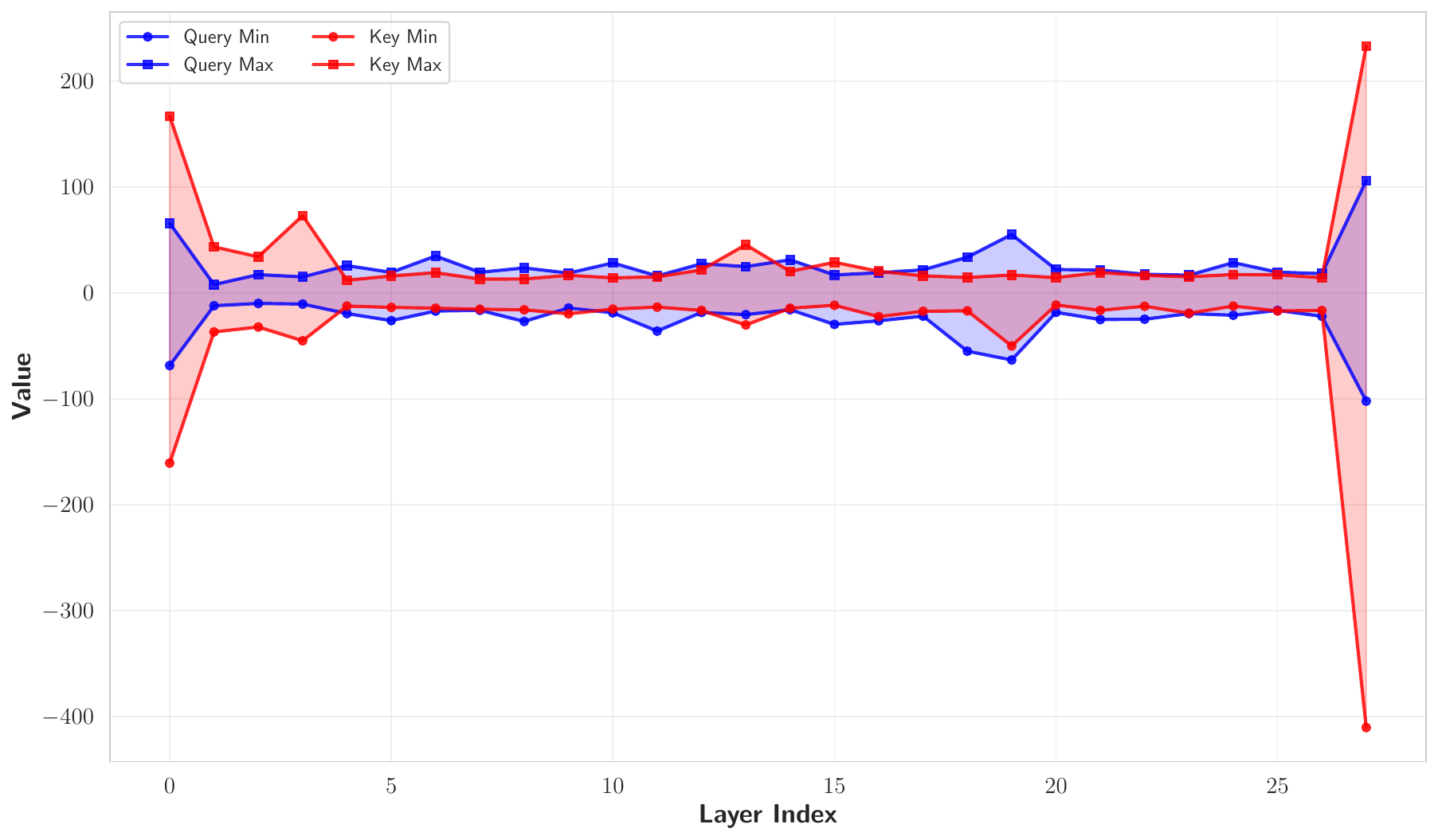}
    \caption{Analysis of numerical magnitude divergence}
    \label{fig:magnitude_divergence}
  \end{minipage}
\end{figure*}

As shown in Figure \ref{fig:scalability}, the scalability comparison reveals fundamental distinctions between the two approaches.

\subsection{Layer Sensitivity and Deployment Strategy}
In the following, we analyze the sensitivity of individual layers to DASH-KV replacement and use the resulting observations to derive an effective deployment strategy.

\subsubsection{Layer-wise Sensitivity Analysis}

We observe that Transformer layers demonstrate significantly varying tolerances to sparsification. To identify the optimal deployment strategy for DASH-KV, we conduct a systematic layer-wise sensitivity analysis. Specifically, we sequentially replace each layer in the Qwen2-7B model with DASH-KV and assess the resulting performance degradation across six LongBench tasks. As illustrated in Figure~\ref{fig:layer_sensitivity}, the sensitivity profile follows an Inverted-U-shaped curve, which is readily interpretable.

\textbf{Early and Later Layers (e.g., Layer 0-4, 27-32).} Replacing these layers with DASH-KV results in a severe performance collapse, as they play a pivotal role in preserving the model’s fundamental features and final representations.

\textbf{Middle Layers (e.g., Layer 5-27).} In sharp contrast, the intermediate layers exhibit considerable robustness to sparsification. Substituting these layers with DASH-KV leads to only marginal performance degradation, indicating that they possess substantial redundancy, which can be effectively compressed without compromising model efficacy.

\subsubsection{Sandwich Deployment Strategy}

Based on the layer-wise sensitivity analysis presented earlier, this paper introduces a Sandwich Deployment Strategy. This approach maintains Full Attention in both the initial and final layers of the model to preserve representational stability, while implementing DASH-KV within the intermediate layers to optimize computational efficiency.

\subsection{Analysis of the rationality of asymmetric hashing.}

\begin{figure}[htbp]
  \centering
  \includegraphics[width=\columnwidth]{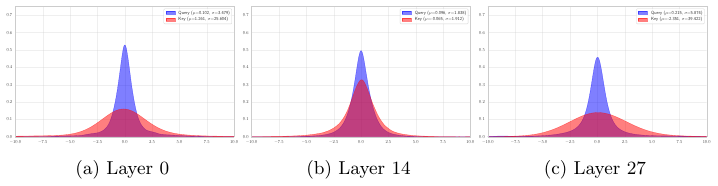}
  \caption{Visualization of distributional characteristics of Queries and Keys across layers}
  \label{fig:distribution_manifold}
\end{figure}

Empirical evidence presented in Figures 8 and 9 demonstrates a marked divergence in both the numerical magnitudes and distributional characteristics between the Query ($Q$) and Key ($K$) representations. This observed discrepancy substantiates the rationale underlying the proposed asymmetric hashing framework. As a symmetric methodology would be inadequate to accommodate the distinct geometric manifolds and value ranges inherent to these two functional components.

\section{Conclusion}

This paper presents DASH-KV, a novel inference acceleration framework leveraging deep hashing retrieval. It seeks to fundamentally address efficiency bottlenecks in long-context reasoning by reconfiguring the underlying computational paradigm. The framework reconstructs the attention mechanism into an approximate nearest neighbor search within Hashing space. This design effectively replaces costly floating-point matrix multiplications with efficient bitwise operations. Comprehensive experiments conducted across multiple long-text benchmarks indicate that DASH-KV attains a superior balance between efficiency and accuracy. The work not only validates the feasibility of integrating hash retrieval into the Transformer architecture but also offers a new technical pathway for the efficient deployment of large-scale models. To further realize the potential of the proposed hashing strategy, future work will involve the development of custom GPU kernels optimized for low-precision bitwise similarity search. By exploiting specialized hardware instructions and refining memory access patterns, we anticipate substantial reductions in the I/O overhead associated with KV cache management, particularly in extreme long-context scenarios.

\section{Limitations}
Although DASH-KV significantly improves efficiency and maintains accuracy in long-context reasoning, our work has certain limitations. In our experiments, we used FP16 format to simulate binary hash code storage and computation. This choice was made for prototyping convenience and to make full use of existing deep learning frameworks' efficient matrix computation kernels. We did not implement low-level bitwise operators directly. However, all reported storage usage and speed-up benefits are based on a strict theoretical 1-bit packed storage model. Experiments have also confirmed that the hash retrieval mechanism performs consistently well. Future native hardware or operator support could fully unlock its efficiency potential.

\section*{Acknowledgements}
This research is supported by Sichuan Science and Technology Program (Grant No. 2026NSFSC1474). This research is also supported by the Postdoctoral Fellowship Program (Grade C) of the China Postdoctoral Science Foundation (Grant No. GZC20251053). This work is partially supported by UESTC Kunpeng\&Ascend Center of Cultivation (Project ID: H04W241592) and the National Natural Science Foundation of China under grant 62572104

\bibliography{custom}

@article{achiam2023gpt,
  title={Gpt-4 technical report},
  author={Achiam, Josh and Adler, Steven and Agarwal, Sandhini and Ahmad, Lama and Akkaya, Ilge and Aleman, Florencia Leoni and Almeida, Diogo and Altenschmidt, Janko and Altman, Sam and Anadkat, Shyamal and others},
  journal={arXiv preprint arXiv:2303.08774},
  year={2023}
}

@article{touvron2023llama,
  title={Llama: Open and efficient foundation language models},
  author={Touvron, Hugo and Lavril, Thibaut and Izacard, Gautier and Martinet, Xavier and Lachaux, Marie-Anne and Lacroix, Timoth{\'e}e and Rozi{\`e}re, Baptiste and Goyal, Naman and Hambro, Eric and Azhar, Faisal and others},
  journal={arXiv preprint arXiv:2302.13971},
  year={2023}
}

@article{pope2023efficiently,
  title={Efficiently scaling transformer inference},
  author={Pope, Reiner and Douglas, Sholto and Chowdhery, Aakanksha and Devlin, Jacob and Bradbury, James and Heek, Jonathan and Xiao, Kefan and Agrawal, Shivani and Dean, Jeff},
  journal={Proceedings of machine learning and systems},
  volume={5},
  pages={606--624},
  year={2023}
}

@article{vyas2020fast,
  title={Fast transformers with clustered attention},
  author={Vyas, Apoorv and Katharopoulos, Angelos and Fleuret, Fran{\c{c}}ois},
  journal={Advances in Neural Information Processing Systems},
  volume={33},
  pages={21665--21674},
  year={2020}
}

@inproceedings{liu2012supervised,
  title={Supervised hashing with kernels},
  author={Liu, Wei and Wang, Jun and Ji, Rongrong and Jiang, Yu-Gang and Chang, Shih-Fu},
  booktitle={2012 IEEE conference on computer vision and pattern recognition},
  pages={2074--2081},
  year={2012},
  organization={IEEE}
}

@inproceedings{lai2015simultaneous,
  title={Simultaneous feature learning and hash coding with deep neural networks},
  author={Lai, Hanjiang and Pan, Yan and Liu, Ye and Yan, Shuicheng},
  booktitle={Proceedings of the IEEE conference on computer vision and pattern recognition},
  pages={3270--3278},
  year={2015}
}

@article{sharma2023truth,
  title={The truth is in there: Improving reasoning in language models with layer-selective rank reduction},
  author={Sharma, Pratyusha and Ash, Jordan T and Misra, Dipendra},
  journal={arXiv preprint arXiv:2312.13558},
  year={2023}
}

@article{chen2021scatterbrain,
  title={Scatterbrain: Unifying sparse and low-rank attention},
  author={Chen, Beidi and Dao, Tri and Winsor, Eric and Song, Zhao and Rudra, Atri and R{\'e}, Christopher},
  journal={Advances in Neural Information Processing Systems},
  volume={34},
  pages={17413--17426},
  year={2021}
}

@article{zhao2024atom,
  title={Atom: Low-bit quantization for efficient and accurate llm serving},
  author={Zhao, Yilong and Lin, Chien-Yu and Zhu, Kan and Ye, Zihao and Chen, Lequn and Zheng, Size and Ceze, Luis and Krishnamurthy, Arvind and Chen, Tianqi and Kasikci, Baris},
  journal={Proceedings of Machine Learning and Systems},
  volume={6},
  pages={196--209},
  year={2024}
}

@article{liu2024kivi,
  title={Kivi: A tuning-free asymmetric 2bit quantization for kv cache},
  author={Liu, Zirui and Yuan, Jiayi and Jin, Hongye and Zhong, Shaochen and Xu, Zhaozhuo and Braverman, Vladimir and Chen, Beidi and Hu, Xia},
  journal={arXiv preprint arXiv:2402.02750},
  year={2024}
}

@article{xiao2023efficient,
  title={Efficient streaming language models with attention sinks},
  author={Xiao, Guangxuan and Tian, Yuandong and Chen, Beidi and Han, Song and Lewis, Mike},
  journal={arXiv preprint arXiv:2309.17453},
  year={2023}
}

@article{zhang2023h2o,
  title={H2o: Heavy-hitter oracle for efficient generative inference of large language models},
  author={Zhang, Zhenyu and Sheng, Ying and Zhou, Tianyi and Chen, Tianlong and Zheng, Lianmin and Cai, Ruisi and Song, Zhao and Tian, Yuandong and R{\'e}, Christopher and Barrett, Clark and others},
  journal={Advances in Neural Information Processing Systems},
  volume={36},
  pages={34661--34710},
  year={2023}
}

@article{li2024snapkv,
  title={Snapkv: Llm knows what you are looking for before generation},
  author={Li, Yuhong and Huang, Yingbing and Yang, Bowen and Venkitesh, Bharat and Locatelli, Acyr and Ye, Hanchen and Cai, Tianle and Lewis, Patrick and Chen, Deming},
  journal={Advances in Neural Information Processing Systems},
  volume={37},
  pages={22947--22970},
  year={2024}
}

@inproceedings{ainslie2023gqa,
  title={Gqa: Training generalized multi-query transformer models from multi-head checkpoints},
  author={Ainslie, Joshua and Lee-Thorp, James and De Jong, Michiel and Zemlyanskiy, Yury and Lebr{\'o}n, Federico and Sanghai, Sumit},
  booktitle={Proceedings of the 2023 Conference on Empirical Methods in Natural Language Processing},
  pages={4895--4901},
  year={2023}
}

@inproceedings{kwon2023efficient,
  title={Efficient memory management for large language model serving with pagedattention},
  author={Kwon, Woosuk and Li, Zhuohan and Zhuang, Siyuan and Sheng, Ying and Zheng, Lianmin and Yu, Cody Hao and Gonzalez, Joseph and Zhang, Hao and Stoica, Ion},
  booktitle={Proceedings of the 29th symposium on operating systems principles},
  pages={611--626},
  year={2023}
}

@article{jegou2010product,
  title={Product quantization for nearest neighbor search},
  author={Jegou, Herve and Douze, Matthijs and Schmid, Cordelia},
  journal={IEEE transactions on pattern analysis and machine intelligence},
  volume={33},
  number={1},
  pages={117--128},
  year={2010},
  publisher={IEEE}
}

@article{andoni2008near,
  title={Near-optimal hashing algorithms for approximate nearest neighbor in high dimensions},
  author={Andoni, Alexandr and Indyk, Piotr},
  journal={Communications of the ACM},
  volume={51},
  number={1},
  pages={117--122},
  year={2008},
  publisher={ACM New York, NY, USA}
}

@inproceedings{guo2016robust,
  title={Robust iterative quantization for efficient lp-norm similarity search},
  author={Guo, Yuchen and Ding, Guiguang and Han, Jungong and Jin, Xiaoming},
  booktitle={Proceedings of the twenty-fifth international joint conference on artificial intelligence},
  pages={3382--3388},
  year={2016}
}

@inproceedings{liu2016deep,
  title={Deep supervised hashing for fast image retrieval},
  author={Liu, Haomiao and Wang, Ruiping and Shan, Shiguang and Chen, Xilin},
  booktitle={Proceedings of the IEEE conference on computer vision and pattern recognition},
  pages={2064--2072},
  year={2016}
}

@inproceedings{bai2024longbench,
  title={Longbench: A bilingual, multitask benchmark for long context understanding},
  author={Bai, Yushi and Lv, Xin and Zhang, Jiajie and Lyu, Hongchang and Tang, Jiankai and Huang, Zhidian and Du, Zhengxiao and Liu, Xiao and Zeng, Aohan and Hou, Lei and others},
  booktitle={Proceedings of the 62nd annual meeting of the association for computational linguistics (volume 1: Long papers)},
  pages={3119--3137},
  year={2024}
}

@article{dao2022flashattention,
  title={Flashattention: Fast and memory-efficient exact attention with io-awareness},
  author={Dao, Tri and Fu, Dan and Ermon, Stefano and Rudra, Atri and R{\'e}, Christopher},
  journal={Advances in neural information processing systems},
  volume={35},
  pages={16344--16359},
  year={2022}
}

@article{ge2023model,
  title={Model tells you what to discard: Adaptive kv cache compression for llms},
  author={Ge, Suyu and Zhang, Yunan and Liu, Liyuan and Zhang, Minjia and Han, Jiawei and Gao, Jianfeng},
  journal={arXiv preprint arXiv:2310.01801},
  year={2023}
}

@article{liu2023scissorhands,
  title={Scissorhands: Exploiting the persistence of importance hypothesis for llm kv cache compression at test time},
  author={Liu, Zichang and Desai, Aditya and Liao, Fangshuo and Wang, Weitao and Xie, Victor and Xu, Zhaozhuo and Kyrillidis, Anastasios and Shrivastava, Anshumali},
  journal={Advances in Neural Information Processing Systems},
  volume={36},
  pages={52342--52364},
  year={2023}
}

@article{kitaev2020reformer,
  title={Reformer: The efficient transformer},
  author={Kitaev, Nikita and Kaiser, {\L}ukasz and Levskaya, Anselm},
  journal={arXiv preprint arXiv:2001.04451},
  year={2020}
}

@inproceedings{cao2017hashnet,
  title={Hashnet: Deep learning to hash by continuation},
  author={Cao, Zhangjie and Long, Mingsheng and Wang, Jianmin and Yu, Philip S},
  booktitle={Proceedings of the IEEE international conference on computer vision},
  pages={5608--5617},
  year={2017}
}

@article{zhou2024survey,
  title={A survey on efficient inference for large language models},
  author={Zhou, Zixuan and Ning, Xuefei and Hong, Ke and Fu, Tianyu and Xu, Jiaming and Li, Shiyao and Lou, Yuming and Wang, Luning and Yuan, Zhihang and Li, Xiuhong and others},
  journal={arXiv preprint arXiv:2404.14294},
  year={2024}
}

@inproceedings{tay2020sparse,
  title={Sparse sinkhorn attention},
  author={Tay, Yi and Bahri, Dara and Yang, Liu and Metzler, Donald and Juan, Da-Cheng},
  booktitle={International conference on machine learning},
  pages={9438--9447},
  year={2020},
  organization={PMLR}
}

@inproceedings{voita2019analyzing,
  title={Analyzing multi-head self-attention: Specialized heads do the heavy lifting, the rest can be pruned},
  author={Voita, Elena and Talbot, David and Moiseev, Fedor and Sennrich, Rico and Titov, Ivan},
  booktitle={Proceedings of the 57th annual meeting of the association for computational linguistics},
  pages={5797--5808},
  year={2019}
}

@article{michel2019sixteen,
  title={Are sixteen heads really better than one?},
  author={Michel, Paul and Levy, Omer and Neubig, Graham},
  journal={Advances in neural information processing systems},
  volume={32},
  year={2019}
}

@article{goyal2017accurate,
  title={Accurate, large minibatch sgd: Training imagenet in 1 hour},
  author={Goyal, Priya and Doll{\'a}r, Piotr and Girshick, Ross and Noordhuis, Pieter and Wesolowski, Lukasz and Kyrola, Aapo and Tulloch, Andrew and Jia, Yangqing and He, Kaiming},
  journal={arXiv preprint arXiv:1706.02677},
  year={2017}
}

@article{hinton2015distilling,
  title={Distilling the knowledge in a neural network},
  author={Hinton, Geoffrey and Vinyals, Oriol and Dean, Jeff},
  journal={arXiv preprint arXiv:1503.02531},
  year={2015}
}

@inproceedings{zhu2016deep,
  title={Deep hashing network for efficient similarity retrieval},
  author={Zhu, Han and Long, Mingsheng and Wang, Jianmin and Cao, Yue},
  booktitle={Proceedings of the AAAI conference on Artificial Intelligence},
  volume={30},
  number={1},
  year={2016}
}

@inproceedings{xiao2023smoothquant,
  title={Smoothquant: Accurate and efficient post-training quantization for large language models},
  author={Xiao, Guangxuan and Lin, Ji and Seznec, Mickael and Wu, Hao and Demouth, Julien and Han, Song},
  booktitle={International conference on machine learning},
  pages={38087--38099},
  year={2023},
  organization={PMLR}
}

@article{tay2022efficient,
  title={Efficient transformers: A survey},
  author={Tay, Yi and Dehghani, Mostafa and Bahri, Dara and Metzler, Donald},
  journal={ACM Computing Surveys},
  volume={55},
  number={6},
  pages={1--28},
  year={2022},
  publisher={ACM New York, NY}
}

@inproceedings{rastegari2016xnor,
  title={Xnor-net: Imagenet classification using binary convolutional neural networks},
  author={Rastegari, Mohammad and Ordonez, Vicente and Redmon, Joseph and Farhadi, Ali},
  booktitle={European conference on computer vision},
  pages={525--542},
  year={2016},
  organization={Springer}
}

@article{wang2017survey,
  title={A survey on learning to hash},
  author={Wang, Jingdong and Zhang, Ting and Sebe, Nicu and Shen, Heng Tao and others},
  journal={IEEE transactions on pattern analysis and machine intelligence},
  volume={40},
  number={4},
  pages={769--790},
  year={2017},
  publisher={IEEE}
}

@inproceedings{andoni2015optimal,
  title={Optimal data-dependent hashing for approximate near neighbors},
  author={Andoni, Alexandr and Razenshteyn, Ilya},
  booktitle={Proceedings of the forty-seventh annual ACM symposium on Theory of computing},
  pages={793--801},
  year={2015}
}

@article{hooper2024kvquant,
  title={Kvquant: Towards 10 million context length llm inference with kv cache quantization},
  author={Hooper, Coleman and Kim, Sehoon and Mohammadzadeh, Hiva and Mahoney, Michael W and Shao, Yakun S and Keutzer, Kurt and Gholami, Amir},
  journal={Advances in Neural Information Processing Systems},
  volume={37},
  pages={1270--1303},
  year={2024}
}

@article{zheng2026llava,
  title={LLaVA-FA: Learning Fourier Approximation for Compressing Large Multimodal Models},
  author={Zheng, Pengcheng and Zhang, Chaoning and Mo, Jiarong and Li, GuoHui and Zhang, Jiaquan and Zhang, Jiahao and Cao, Sihan and Zheng, Sheng and Qin, Caiyan and Wang, Guoqing and others},
  journal={arXiv preprint arXiv:2602.00135},
  year={2026}
}

@article{zheng2026towards,
  title={Towards visual chain-of-thought reasoning: A comprehensive survey},
  author={Zheng, Pengcheng and Zhang, Chaoning and Cui, Mingzheng and Chen, Guo and Sun, Qigan and Huang, Jiaxin and Zhang, Jiaquan and Kim, Tae-Ho and Qin, Caiyan and Ren, Yazhou and others},
  year={2026},
  publisher={TechRxiv}
}

@article{cao2026language,
  title={Language-guided token compression with reinforcement learning in large vision-language models},
  author={Cao, Sihan and Zhang, Jianwei and Zheng, Pengcheng and Yan, Jiaxin and Qin, Caiyan and Ye, Yalan and Dong, Wei and Wang, Peng and Yang, Yang and Zhang, Chaoning},
  journal={arXiv preprint arXiv:2603.13394},
  year={2026}
}

@article{zheng2025joint,
  title={Joint lossless compression and steganography for medical images via large language models},
  author={Zheng, Pengcheng and Pu, Xiaorong and Chen, Kecheng and Huang, Jiaxin and Yang, Meng and Feng, Bai and Ren, Yazhou and Jiang, Jianan and Zhang, Chaoning and Yang, Yang and others},
  journal={arXiv preprint arXiv:2508.01782},
  year={2025}
}

@article{sun2026grasp,
  title={GRASP: Guided Region-Aware Sparse Prompting for Adapting MLLMs to Remote Sensing},
  author={Sun, Qigan and Zhang, Chaoning and Zhang, Jianwei and Wang, Xudong and Xie, Jiehui and Zheng, Pengcheng and Wang, Haoyu and Lee, Sungyoung and Tai, Chi-lok Andy and Yang, Yang and others},
  journal={arXiv preprint arXiv:2601.17089},
  year={2026}
}

@article{wang2026efficient,
  title={Efficient and Interpretable Multi-Agent LLM Routing via Ant Colony Optimization},
  author={Wang, Xudong and Zhang, Chaoning and Zhang, Jiaquan and Li, Chenghao and Sun, Qigan and Bae, Sung-Ho and Wang, Peng and Xie, Ning and Zou, Jie and Yang, Yang and others},
  journal={arXiv preprint arXiv:2603.12933},
  year={2026}
}

@article{wang2026transforming,
  title={Transforming external knowledge into triplets for enhanced retrieval in rag of llms},
  author={Wang, Xudong and Zhang, Chaoning and Sun, Qigan and Huang, Zhenzhen and Lu, Chang and Zheng, Sheng and Ma, Zeyu and Qin, Caiyan and Yang, Yang and Shen, Hengtao},
  journal={arXiv preprint arXiv:2604.12610},
  year={2026}
}

@article{yuan2025riemannian,
  title={Riemannian optimization on relaxed indicator matrix manifold},
  author={Yuan, Jinghui and Xie, Fangyuan and Nie, Feiping and Li, Xuelong},
  journal={arXiv preprint arXiv:2503.20505},
  year={2025}
}

@inproceedings{yuan2026spherical,
  title={Spherical cautious optimizers},
  author={Yuan, Jh and Nie, Feiping},
  booktitle={Workshop on Scientific Methods for Understanding Deep Learning},
  year={2026}
}

@article{zhang2026learning,
  title={Learning global hypothesis space for enhancing synergistic reasoning chain},
  author={Zhang, Jiaquan and Zhang, Chaoning and Chen, Shuxu and Wang, Xudong and Huang, Zhenzhen and Zheng, Pengcheng and Yuan, Shuai and Zheng, Sheng and Sun, Qigan and Zou, Jie and others},
  journal={arXiv preprint arXiv:2602.09794},
  year={2026}
}

@article{zhang2026text,
  title={Text summarization via global structure awareness},
  author={Zhang, Jiaquan and Zhang, Chaoning and Chen, Shuxu and Liu, Yibei and Li, Chenghao and Sun, Qigan and Yuan, Shuai and Puspitasari, Fachrina Dewi and Han, Dongshen and Wang, Guoqing and others},
  journal={arXiv preprint arXiv:2602.09821},
  year={2026}
}

@article{zhang2026lightweight,
  title={Lightweight llm agent memory with small language models},
  author={Zhang, Jiaquan and Zhang, Chaoning and Chen, Shuxu and Huang, Zhenzhen and Zheng, Pengcheng and Wang, Zhicheng and Guo, Ping and Mo, Fan and Bae, Sung-Ho and Zou, Jie and others},
  journal={arXiv preprint arXiv:2604.07798},
  year={2026}
}

@article{zhang2026tda,
  title={TDA-RC: Task-Driven Alignment for Knowledge-Based Reasoning Chains in Large Language Models},
  author={Zhang, Jiaquan and Sun, Qigan and Zhang, Chaoning and Wang, Xudong and Huang, Zhenzhen and Zhou, Yitian and Zheng, Pengcheng and Tai, Chi-lok Andy and Bae, Sung-Ho and Ma, Zeyu and others},
  journal={arXiv preprint arXiv:2604.04942},
  year={2026}
}

\appendix

\section{Additional Related Work}
\label{sec:appendix}
\subsection{KV Cache Optimization for Long-Context Inference}

To alleviate the bottleneck of KV Cache in long-context inference, existing works primarily follow three trajectories. First, quantization compression (e.g., KIVI, Atom) reduces memory usage through low-bit representation. However, this approach suffers from significant accuracy loss at extremely low bit-widths and introduces additional dequantization overhead. Second, selective caching (e.g., StreamingLLM, H2O) eliminates redundancy based on attention sparsity. Although this reduces memory usage to a constant level, rigid eviction strategies result in the permanent loss of historical information, thereby compromising long-range dependency performance. Third, structured sharing (e.g., GQA, PyramidKV) reduces redundancy via cross-head or cross-layer sharing. Nevertheless, these methods often overlook fine-grained internal model heterogeneity and typically require integration during the pre-training phase, making flexible application to existing models difficult.

While the aforementioned methods achieve significant storage compression, they share a common limitation: they fail to fundamentally resolve the attention computation bottleneck. Whether through quantization or sparsification, the core computation still relies on similarity matching of high-dimensional floating-point vectors. Consequently, the computational complexity remains bound by the floating-point operation paradigm, failing to reach the theoretical upper limit of computational efficiency.

\subsection{Deep Hashing Methods}

Hashing techniques map high-dimensional vectors to binary codes, utilizing bitwise operations to achieve low-latency and low-memory Approximate Nearest Neighbor (ANN) search. Unlike tree-based, quantization-based, or graph-based methods, hashing focuses on preserving bit-level locality. Early random projection methods, represented by Locality-Sensitive Hashing (LSH), were simple but suffered from limited recall rates. Subsequently, learning-based hashing methods (such as Spectral Hashing and ITQ) improved accuracy with shorter code lengths by optimizing mapping functions. In recent years, deep supervised hashing (e.g., DSH, CNNH) has become mainstream by leveraging neural networks to capture complex semantics. Furthermore, asymmetric hashing and multi-hash ensemble strategies have enhanced adaptability in large-scale scenarios like Retrieval-Augmented Generation (RAG) by reducing quantization errors and employing multi-view fusion.

In summary, while approximate retrieval technologies represented by deep hashing have achieved significant success and maturity in fields such as large-scale image retrieval, few have noted the high degree of isomorphism between their technical paradigms and critical scenarios like Transformer internal attention computation. To the best of our knowledge, this work is the first to systematically introduce advanced large-scale data retrieval techniques into the internal optimization of the Transformer attention mechanism. This approach aims to break the traditional paradigm bottleneck dominated by large-scale, high-dimensional floating-point vector similarity calculations during autoregressive generation, thereby opening new technical pathways for next-generation efficient Large Language Model (LLM) inference.

\section{Datasets}
\label{appendix:datasets}
We evaluate our method on six representative tasks from the LongBench benchmark \cite{bai2024longbench}, selected to cover a diverse range of long-context capabilities including single-document QA, multi-document summarization, and multi-hop reasoning. The specifications of each dataset are detailed below:

\paragraph{NarrativeQA}
This dataset targets reading comprehension over long narratives, such as book summaries or movie scripts. It requires the model to understand the underlying plot and causal relationships within extensive texts. The task involves answering specific questions based on a single long document, challenging the model's ability to maintain context over long sequences.

\paragraph{HotpotQA}
Focused on multi-hop reasoning, HotpotQA requires models to synthesize information from multiple supporting documents to answer a question. Unlike simple retrieval tasks, the model must bridge distinct pieces of evidence to derive the correct answer. This tests the ability to handle fragmented information and perform logical deduction across multiple contexts.

\paragraph{Qasper}
This dataset consists of question-answering tasks based on academic research papers from the NLP domain. It evaluates the model's capacity to extract technical details, summarize specific sections, and interpret experimental results. The inputs typically include full-text papers, requiring the processing of dense, domain-specific information.

\paragraph{MultiNews}
Designed for multi-document summarization, MultiNews requires the model to generate a concise summary based on several news articles covering the same event. This task assesses the ability to identify common themes, resolve conflicting information, and synthesize a coherent narrative from diverse sources.

\paragraph{GovReport}
This dataset involves summarizing long government reports. The source documents contain formal, policy-oriented language with complex structures. The task challenges the model to distill key policy details and hierarchical information into a comprehensive summary, testing performance on highly structured and formal long-context data.

\paragraph{TriviaQA}
TriviaQA is a reading comprehension dataset containing question-answer pairs accompanied by evidence documents. The task requires the model to locate factual answers within large amounts of noisy text. It evaluates the efficacy of information retrieval and the precision of factual extraction in long-context scenarios.

\section{Models}
We employ three distinct Large Language Models (LLMs) to evaluate the versatility and scalability of DASH-KV. The specific details of each model are as follows:

\paragraph{Qwen2-7B-Instruct}
Developed by Alibaba Cloud, this model represents the highly optimized 7B parameter class. It utilizes Grouped Query Attention (GQA) in its architecture. GQA significantly reduces the memory footprint of the KV cache during inference. This model serves as a standard baseline for evaluating efficiency in compact dense models.

\paragraph{Llama-3.1-8B-Instruct}
This model is the latest iteration from Meta, featuring native support for a context window of up to 128k tokens. It employs optimized Rotary Positional Embeddings (RoPE) to handle long sequences effectively. The inclusion of Llama-3.1 allows us to assess the robustness of DASH-KV in scenarios requiring ultra-long context retention.

\paragraph{Qwen2.5-14B-Instruct}
This model introduces a larger parameter scale compared to the 7B and 8B baselines. It benefits from extensive pre-training on a diverse corpus, resulting in superior reasoning capabilities. Testing on this model verifies whether the performance gains of DASH-KV remain consistent as model size and architectural complexity increase.

\section{Theoretical Equivalence of Bitwise Operations}
\label{sec:theory_equivalence}
A critical premise of our approach is that the computationally expensive floating-point similarity search can be mathematically equivalently replaced by efficient bitwise operations. In our experiments, we simulate the retrieval process using matrix multiplication for ease of implementation within deep learning frameworks. Here, we provide the theoretical proof that this simulation is strictly equivalent to the Hamming distance computation in the binary space.

Consider the query hash code $h_q \in \{-1, +1\}^D$ and the key hash code $h_k \in \{-1, +1\}^D$, where $D$ denotes the embedding dimension (e.g., 128). 

The \textbf{Inner Product} in the binary space is defined as:
\begin{equation}
\text{Sim}(h_q, h_k) = \sum_{i=1}^{D} h_q^{(i)} \cdot h_k^{(i)}
\end{equation}
Since $h_q^{(i)}, h_k^{(i)} \in \{-1, +1\}$, the product term $h_q^{(i)} \cdot h_k^{(i)}$ takes the value $+1$ if the bits are identical ($h_q^{(i)} = h_k^{(i)}$) and $-1$ if they are different ($h_q^{(i)} \neq h_k^{(i)}$).

The \textbf{Hamming Distance} $H(h_q, h_k)$ counts the number of positions where the bits differ. Let $N_{diff}$ be the number of differing bits (Hamming distance) and $N_{same}$ be the number of identical bits. We have the following relationships:
\begin{align}
N_{diff} &= H(h_q, h_k) \\
N_{same} &= D - H(h_q, h_k)
\end{align}
Substituting these into the inner product formula:
\begin{align}
\text{Sim}(h_q, h_k) &= N_{same} \cdot (+1) + N_{diff} \cdot (-1) \\
&= (D - H(h_q, h_k)) - H(h_q, h_k) \\
&= D - 2 \cdot H(h_q, h_k)
\end{align}

This derivation proves a strict linear mapping between the inner product and the Hamming distance:
\begin{equation}
\label{eq:equivalence}
\arg\max_{k} \text{Sim}(h_q, h_k) \iff \arg\min_{k} H(h_q, h_k)
\end{equation}

In our PyTorch-based experiments, we compute similarity using $\text{sign}(\mathbf{Q}) \cdot \text{sign}(\mathbf{K})^\top$ (floating-point matrix multiplication on $\{-1, +1\}$ tensors). Equation~\ref{eq:equivalence} guarantees that the ranking of candidates obtained via this simulation is identical to that obtained via hardware-level bitwise operations (XOR followed by POPCNT). Therefore, the accuracy metrics reported in our experiments faithfully reflect the performance of the proposed bitwise acceleration paradigm.

\section{Introduction of residuals}
\subsection{The necessity of residuals}
To elucidate the necessity of the residual MLP, we address a fundamental limitation of binary hashing within the Transformer context: Magnitude-Blindness. Statistics on Key vectors reveal an immense dynamic range (e.g., spanning $[- -412, 236]$ in Layer 27) accompanied by significant outliers. Binary hashing compresses both negligible activations and extreme values into identical codewords. This results in "magnitude collapse," hindering the identification of critical "Attention Sinks." The residual MLP remedies this by leveraging continuous representations to detect high-activation features and predict magnitude compensations. By effectively recovering the attention intensity lost during binarization, this approach significantly outperforms pure hashing in scenarios involving heavy-tailed distributions.

Unlike discrete hashing, the MLP exploits continuous space properties to capture original magnitudes, allowing it to predict substantial compensation for high-activation outliers. This effective restoration of the attention intensity lost to binarization establishes the significant superiority of DASH-KV over pure hashing in layers characterized by heavy-tailed distributions.

\subsection{Analysis of the selection of the necessary residual module loss function}
We evaluated three progressive variants to quantify the impact of residual learning: Pure Hash (without residual compensation), DASH-KV-MSE (trained via Mean Squared Error), and DASH-KV-Distill (trained via distillation).

The experimental results reveal a critical insight: directly fitting the residual function using Mean Squared Error (MSE) results in a significant performance degradation (F1 score drops by 0.39, from 11.31 to 10.92). The underlying cause of this phenomenon is the Scale Shift during Length Extrapolation. When the model learns residuals using the MSE objective on short sequences (3k tokens), it overfits to point-wise absolute value predictions. However, when extrapolated to unseen long contexts (32k tokens) during inference, the numerical range of attention weights shifts significantly, introducing systematic bias into the MSE-optimized residual predictions.

In contrast, the KL divergence-based distillation strategy (Distill) employs a fundamentally different optimization paradigm. Rather than fitting absolute residual values, it guides the residual network to learn the Relative Probability Distribution—specifically, adjusting the relative importance rankings among different Keys rather than their absolute similarity scores. This strategy exhibits inherent robustness for length extrapolation: provided the relative attention rankings remain consistent between training and inference, the model can generalize to arbitrary sequence lengths. Our experiments validate this hypothesis: DASH-KV-Distill effectively mitigates the quantization errors introduced by pure hashing, achieving a +0.13 F1 improvement over the full-precision baseline (11.44 vs. 11.31). This underscores the superiority of the distillation objective in long-context scenarios.

\begin{table*}[t]
\centering
\caption{\textbf{Performance Comparison on Qwen2-7B-Instruct.} We compare the Full Attention baseline (Original) against variants where specific layers are replaced by our method. The metrics reported are F1 scores for NarrativeQA, HotpotQA, Qasper, MultiNews, and GovReport, and Exact Match (EM) for TriviaQA. \textbf{Avg} denotes the average score across all tasks. Note that replacing Layer 0 results in significant degradation, highlighting the sensitivity of early layers.}
\label{tab:qwen_results}
\resizebox{\textwidth}{!}{%
\begin{tabular}{l|cccccc|c}
\toprule
\textbf{Configuration} & \textbf{NarrativeQA} & \textbf{HotpotQA} & \textbf{Qasper} & \textbf{MultiNews} & \textbf{GovReport} & \textbf{TriviaQA} & \textbf{Avg} \\
\midrule
\textbf{Original (Full Attention)} & \textbf{25.13} & 44.04 & \textbf{46.13} & \textbf{15.42} & 18.06 & 83.47 & 38.71 \\
\midrule
Replace Layer 14 & 23.64 & \textbf{44.88} & 44.94 & 15.39 & 18.22 & 83.48 & 38.43 \\
Replace Layer 18 & 24.65 & 44.50 & 45.34 & 15.40 & \textbf{19.13} & 83.33 & \textbf{38.73} \\
Replace Layer 27 & 24.66 & 43.33 & 45.59 & 15.30 & 18.34 & 83.46 & 38.47 \\
Replace Layers 10--24 (Even) & 24.01 & 43.19 & 44.01 & 15.13 & 18.43 & \textbf{84.37} & 38.19 \\
\midrule
\textit{Replace Layer 0 (Ablation)} & \textit{8.74} & \textit{21.74} & \textit{12.40} & \textit{2.62} & \textit{3.10} & \textit{31.22} & \textit{13.30} \\
\bottomrule
\end{tabular}%
}
\end{table*}

\section{The necessity of layer-wise training}

Our empirical investigation uncovers a fundamental challenge in hashing-based attention optimization~\cite{zheng2026towards, zheng2025joint}: the significant distributional heterogeneity of Query ($\mathbf{Q}$) and Key ($\mathbf{K}$) representations across network depths. As widely recognized, Large Language Models process information hierarchically, with shallower layers encoding local syntactic patterns and deeper layers capturing abstract semantic dependencies. This functional distinctiveness suggests that the optimal low-rank manifolds for preserving similarity structures are layer-dependent.

\begin{figure}[t]
    \centering
    \includegraphics[width=\columnwidth]{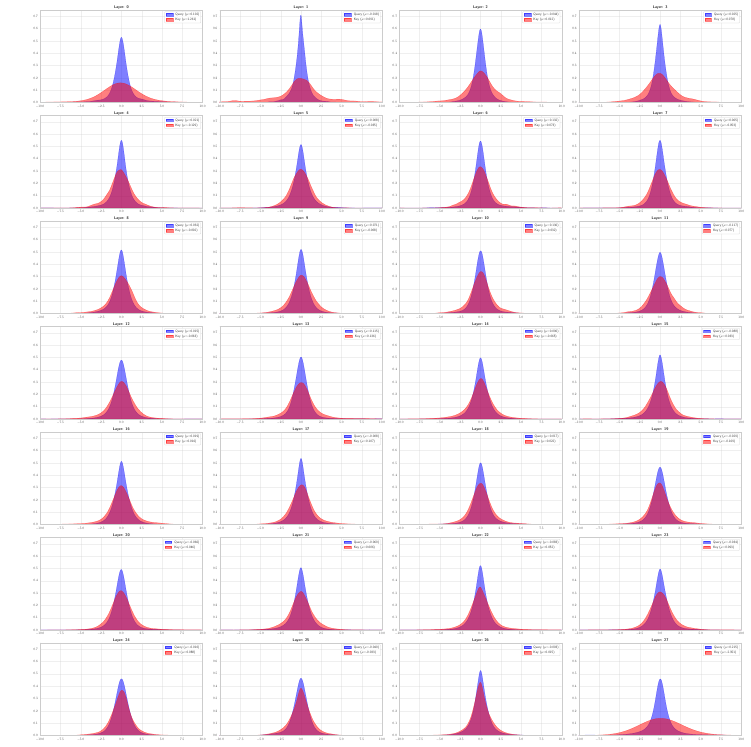}
    \caption{\textbf{Visualization of Query and Key Activation Distributions Across Layers.} We project the Query and Key vectors from representative layers (e.g., Layer 0, 14, 27) into a 2D space. The marked visual discrepancies indicate that the geometric distribution of attention features drifts significantly as network depth increases, implying that a static global hashing projection cannot optimally fit all layers simultaneously.}
    \label{fig:layer_distribution_viz}
\end{figure}

To substantiate this hypothesis, we first visualize the activation landscapes of different layers in Figure~\ref{fig:layer_distribution_viz}. The visualization reveals distinct clustering patterns and density variations, confirming that the feature subspaces undergo substantial transformation (i.e., manifold shift) from bottom to top layers. Consequently, a "one-size-fits-all" hashing projection trained on a specific distribution is theoretically insufficient to capture the diverse geometric characteristics present throughout the model.

\begin{figure}[htbp]
    \centering
    \includegraphics[width=\columnwidth]{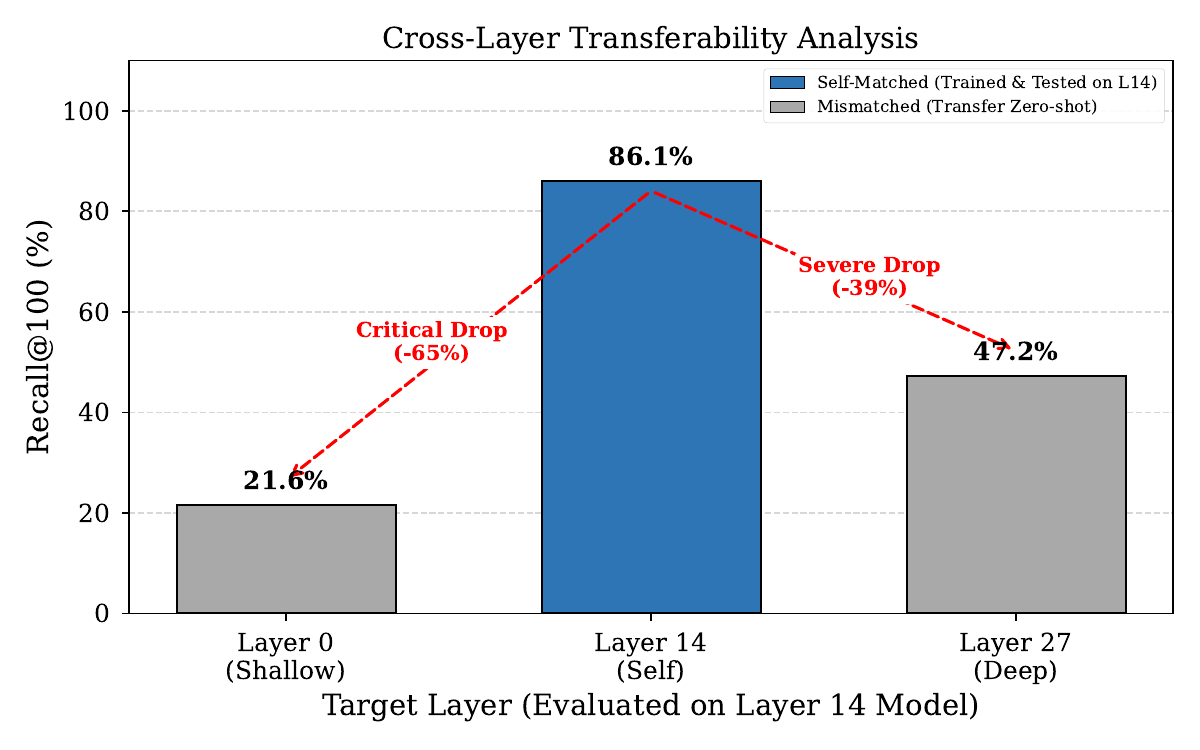}
    \caption{\textbf{Impact of Cross-Layer Weight Transfer on Retrieval Recall.} The figure illustrates the performance degradation when applying hash functions trained on an intermediate layer (Layer 14) to a shallow layer (Layer 1) and a deep layer (Layer 27). The sharp decline in recall confirms the distributional mismatch between layers.}
    \label{fig:layer_wise_transfer}
\end{figure}

To strictly quantify the impact of this heterogeneity, we conducted a cross-layer generalization experiment (see Figure~\ref{fig:layer_wise_transfer}). Specifically, we isolated the hashing network weights optimized on the data distribution of an intermediate layer (Layer 14) and directly transferred them to perform retrieval on a shallow layer (Layer 1) and a deep layer (Layer 27).

As evidenced by the results, this direct weight transfer precipitates a marked degradation in recall performance compared to layer-specific baselines. The inability of Layer 14's parameters to generalize to other depths corroborates our visual findings: the geometric structure of the attention mechanism is highly layer-specific. This phenomenon underscores the limited generalizability of a single global weight configuration. Therefore, we conclude that \textbf{independent parameter optimization for each layer (Layer-Wise Training)} is not merely an engineering choice but a theoretical necessity for ensuring high-fidelity retrieval and minimizing quantization errors across the entire model.

\section{Task Performance}

To rigorously evaluate the efficacy of our proposed method, we conducted extensive experiments using the Qwen2-7B-Instruct model as our primary testbed. We systematically replaced the standard attention mechanism with our optimized module across various layer configurations—ranging from specific single layers to multiple intermediate layers. This ablation study aims to verify whether our approach can maintain high-fidelity retrieval and reasoning capabilities compared to the full-attention baseline.

The quantitative results are summarized in Table~\ref{tab:qwen_results}. Our method demonstrates remarkable robustness across all six long-context benchmarks. Notably, replacing the attention mechanism in deeper layers (e.g., Layer 18 or Layer 27) yields performance highly comparable to the original model. For instance, the configuration replacing Layer 18 achieves an average score of 38.73, which marginally surpasses the full-attention baseline of 38.71.

Furthermore, we explored an aggressive compression strategy by replacing a block of eight intermediate layers (Layers 10, 12, ..., 24). Even under this setting, the model maintains a competitive average score of 38.19, exhibiting negligible degradation in multi-hop reasoning (HotpotQA) and summarization tasks (MultiNews). However, we observed that replacing the very first layer (Layer 0) leads to a catastrophic performance drop (13.30 Avg), underscoring the critical role of initial layers in preserving raw token semantics, which are sensitive to approximation.

\end{document}